\documentclass[a4paper,fleqn]{cas-sc}

\usepackage[authoryear,longnamesfirst]{natbib}

\usepackage{fontawesome5}
\usepackage{graphicx}
\usepackage{amsmath}
\usepackage{amsthm}
\usepackage{amssymb}
\usepackage{amsfonts}
\usepackage{bm}
\usepackage{caption}
\usepackage{multirow}
\usepackage[english]{babel}
\usepackage{adjustbox}
\usepackage{xspace}
\usepackage{subfigure}
\usepackage{booktabs}
\usepackage{enumitem}
\usepackage{color}
\usepackage{colortbl}
\usepackage{textcomp}
\usepackage{float}

\usepackage{array}
\usepackage{algorithm}
\usepackage{listings}
\usepackage{algpseudocode}

\usepackage[frozencache,cachedir=.]{minted}

\usepackage{mdframed}
\usepackage{tcolorbox}
\definecolor{cvprblue}{rgb}{0.21,0.49,0.74}
\usepackage{hyperref}

\setlist[itemize]{topsep=0pt, partopsep=0pt, parsep=0pt, itemsep=0pt}

\usepackage{xcolor}
\definecolor{mylightblue}{RGB}{100,149,237}

\newcommand{\model}{\textsc{UrbanX}\xspace}

\tcbuselibrary{minted,skins, breakable}
\definecolor{myblue}{RGB}{100,149,237} 
\definecolor{mylightblue}{HTML}{E6F0FA}
\tcbset{
  enhanced, 
  colback=white!200!black, 
  colframe=mylightblue, 
  colbacktitle=mylightblue, 
  title filled, 
  coltitle=myblue, 
  fonttitle=\bfseries, 
  arc=3mm, 
  outer arc=3mm, 
  boxrule=0.5mm, 
  toprule=0.5mm, 
  bottomrule=0.5mm, 
  titlerule=0.5mm, 
  drop fuzzy shadow, 
  minted options={ 
    breaklines, 
    fontsize=\footnotesize, 
    linenos, 
    numbersep=2mm, 
    framesep=1.5mm, 
  },
}

\begin{document}
\let\WriteBookmarks\relax
\def\floatpagepagefraction{1}
\def\textpagefraction{.001}

\hypersetup{
  citecolor=cvprblue,
  linkcolor=cvprblue,  
  urlcolor=cvprblue     
}

\shorttitle{UrbanX}

\shortauthors{Yihong Tang et~al.}

\title [mode = title]{From Street Views to Urban Science: Discovering Road Safety Factors with Multimodal Large Language Models}

\author[1]{ Yihong Tang}

\author[2]{ Ao Qu}

\author[3]{ Xujing Yu}

\author[3]{ Weipeng Deng}

\author[3]{ Jun Ma}

\author[2]{ Jinhua Zhao}

\author[1]{ Lijun Sun}

\cormark[1]

\affiliation[1]{organization={McGill University},
    city={Montreal},
    country={Canada}}

\affiliation[2]{organization={Massachusetts Institute of Technology},
    city={Boston},
    country={United States}}

\affiliation[3]{organization={The University of Hong Kong},
    city={Hong Kong SAR},
    country={China}}

\nonumnote{$^\ast$Corresponding author \quad \textsuperscript{\faIcon{github}} Project: \url{https://github.com/YihongT/UrbanX.git}}

\nonumnote{\textsuperscript{\faEnvelopeSquare} \texttt{\href{mailto://yihong.tang@mail.mcgill.ca}{yihong.tang@mail.mcgill.ca}} (Y. Tang) \quad \textsuperscript{\faEnvelopeSquare} \texttt{\href{mailto://lijun.sun@mcgill.ca}{lijun.sun@mcgill.ca}} (L. Sun)\vspace{-5pt}}

\begin{abstract}
\noindent Urban and transportation research has long sought to uncover statistically meaningful relationships between key variables and societal outcomes such as road safety, aiming to generate actionable insights that guide the planning, development, and renewal of urban mobility systems. However, traditional workflows face several key challenges: (1) reliance on human experts to propose hypotheses, which can be time-consuming and prone to confirmation bias; (2) limited interpretability, particularly in deep learning approaches; and (3) underutilization of unstructured data that encodes critical urban context.
To address these limitations, we propose a Multimodal Large Language Model (MLLM)-based approach for interpretable hypothesis inference, enabling the automated generation, assessment, and refinement of hypotheses concerning urban form and transportation safety. 
Specifically, we leverage MLLMs to generate road safety–relevant questions and automatically answer them based on street view images (SVIs) through visual question answering (VQA). These responses are used to construct interpretable embeddings for each SVI, which are then incorporated into linear statistical models for transparent and explainable regression analysis.
\model supports iterative hypothesis testing and refinement guided by statistical evidence, such as coefficient significance, thereby enabling rigorous, transparent scientific discovery of previously overlooked correlations between urban design and transportation risk.
We evaluate our framework on Manhattan street segments and demonstrate that it outperforms pretrained deep learning baselines while offering full interpretability. Beyond road safety, \model can serve as a general-purpose foundation for hypothesis-driven urban mobility analysis, extracting structured insights from unstructured data across diverse socioeconomic and environmental outcomes. 
This approach establishes a scalable and trustworthy pathway for interpretable, data-driven scientific discovery in urban and transportation systems using foundation models such as MLLMs.

\end{abstract}


\maketitle

\section{Introduction}

Understanding how the physical structure of cities shapes societal outcomes is a core objective of urban and transportation science. Researchers across domains such as transportation, planning, and public policy have long sought to uncover statistically meaningful links between the built environment and key outcomes such as traffic safety~\cite{yu2024can,xue2024integrating}, walkability~\cite{ewing2009measuring,ignatius2024digital}, equity~\cite{guzman2017urban,bang2025mobility}, and environmental health~\cite{majchrowska2022deep}. At the heart of this pursuit is scientific discovery, which involves identifying interpretable and generalizable factors that explain urban phenomena and inform data-driven decision-making~\cite{batty2024computable}. However, this process is often challenged by the complexity and heterogeneity of urban form. Much of the relevant contextual information is embedded in unstructured formats such as street-level imagery, architectural layouts, and visual cues tied to human perception~\cite{biljecki2021street}, which remain difficult to quantify through conventional feature engineering or structured data pipelines~\cite{nie2025exploring}.

Despite recent progress in data-driven urban research, existing methodological pipelines face critical challenges in discovering new, interpretable factors from complex urban environments. Traditional approaches often rely on expert-curated variables, black-box predictive models, or handcrafted metrics to study specific urban dimensions~\cite{xia2025reimagining,hu2024does}. These strategies are limited in three fundamental ways. 
First, hypothesis generation is typically manual and cognitively intensive, relying heavily on domain expertise and prior assumptions. This process can be slow and is susceptible to confirmation bias~\cite{gettys1979hypothesis}. Second, while deep learning methods have achieved strong predictive performance, they often operate through latent feature spaces that hinder interpretability and reduce scientific transparency~\cite{lipton2018mythos}. Third, unstructured urban data, particularly street view imagery (SVI), remains an underutilized source of contextual information. Although SVIs contain rich visual cues about the built environment and social space, current models struggle to transform them into structured, meaningful variables~\cite{tang2025large}. 
These limitations make it difficult to explore the urban hypothesis space at scale, especially in transportation safety and mobility applications where transparency, generalizability, and human interpretability are essential for real-world adoption.

\begin{figure}[t]
    \centering
    \includegraphics[width=.7\linewidth]{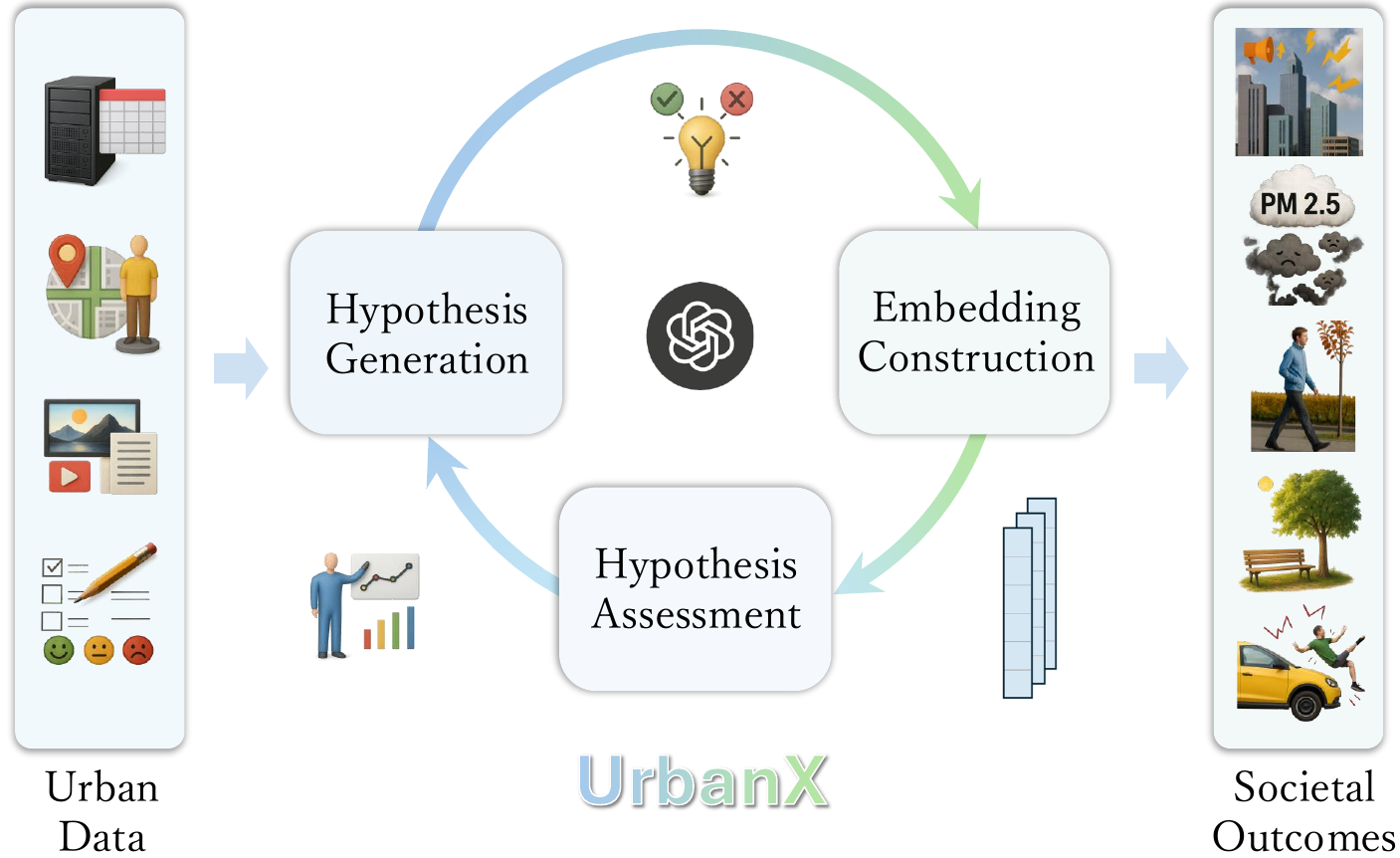}
    \caption{\model: an interpretable, MLLM-powered framework for hypothesis-driven urban scientific discovery.}
    \label{fig:overview}
\end{figure}

Recent advances in foundation models, particularly Large Language Models (LLMs)~\cite{naveed2023comprehensive}, have transformed data-driven reasoning in science and engineering. Trained on large-scale corpora of natural language, LLMs are capable of understanding, generating, and reasoning over text in flexible and context-aware ways~\cite{wei2022emergent}. Building on this progress, their capabilities have been extended to include visual and other modalities, resulting in the development of Multimodal Large Language Models (MLLMs)~\cite{zhang2024mm,wu2023multimodal}. These models jointly process inputs from multiple modalities, including text, images, audio, and video, enabling tasks such as visual question answering, open-ended scene understanding, and multimodal reasoning with minimal supervision.
MLLMs can extract semantically meaningful concepts from visual data and map them to natural language representations that reflect human understanding. In contrast to conventional computer vision models, which primarily focus on perception or classification, MLLMs offer a new capability: generating structured, interpretable variables directly from raw visual inputs. This shift provides a scalable and transparent approach to analyzing unstructured urban data, such as street view imagery, and supports cognitively grounded inference in urban and transportation research.
These capabilities open up new possibilities for interpretable and hypothesis-driven scientific discovery in complex urban mobility systems, where cross-modal reasoning, statistical rigor, and real-world applicability are all essential.

Building on these capabilities, we introduce \model, a framework for interpretable and hypothesis-driven scientific discovery in urban domains, powered by MLLMs. 
As illustrated in Figure~\ref{fig:overview}, \model leverages MLLMs not merely as perception tools but as active collaborators in the scientific reasoning process. Instead of using learned models solely for prediction, our framework treats them as inference engines that support hypothesis generation, extract interpretable variables from structured and unstructured urban data, and evaluate their statistical relevance to real-world societal outcomes.
\model operates through an iterative procedure. At each step, it formulates hypotheses as natural language queries, applies MLLMs to extract semantically meaningful features from urban data, and assesses their explanatory power using interpretable statistical models. Hypotheses with insufficient statistical support are discarded, and new ones are proposed based on observed statistical evidence, allowing the system to gradually converge on a compact set of variables that are both human-aligned and empirically grounded.

We instantiate this framework in the domain of urban road safety, a critical area where interpretability is essential for generating actionable insights and where discovering new, semantically meaningful factors can inform transportation planning and policy. To evaluate its effectiveness, we conduct a case study on street segments in Manhattan, where \model identifies novel variables from SVIs that show significant correlations with crash rates. Our approach achieves better predictive performance than pretrained deep learning baselines such as ResNet~\cite{he2016deep} and Vision Transformer (ViT)~\cite{dosovitskiy2020image}, while maintaining full transparency through interpretable variable construction and attribution.
These results demonstrate the practical utility of \model and support its broader relevance to data-driven urban and transportation research. 
Our work makes the following contributions:
\begin{itemize}[leftmargin=*]
    \item We formalize scientific discovery in urban domains as an inference problem over a hypothesis space, where each hypothesis is a natural language statement linking urban form to societal outcomes. This formulation enables machines to generate, evaluate, and refine hypotheses directly from structured and unstructured data, establishing a scalable foundation for data-driven urban and transportation research.

    \item We introduce a novel use of MLLMs as semantic engines that translate unstructured inputs, such as SVIs, into structured and interpretable variables guided by natural language hypotheses. This approach integrates visual perception with statistical reasoning in a unified and human-interpretable framework, expanding the methodological toolkit for urban and transportation systems analysis.
    
    \item We develop an interpretable, nonparametric framework for hypothesis inference, implemented as an iterative procedure over the hypothesis space. At each iteration, the framework generates new hypotheses, constructs semantically aligned representations, and evaluates variable significance using transparent statistical models. This process enables scalable, statistically grounded discovery of novel urban factors while reducing manual effort in scientific discovery and exploratory analysis.
    
    \item We apply our framework to study road safety in the Manhattan area and demonstrate its ability to uncover novel, interpretable visual variables that significantly correlate with crash rates. Our approach outperforms strong vision baselines in predictive performance while maintaining interpretability through hypothesis-level attribution. Beyond this case study, our approach serves as a general-purpose framework for scientific discovery in urban domains, with the potential to reveal structured insights from unstructured data across a wide range of socioeconomic and environmental outcomes.
\end{itemize}

\section{Related Work}
\label{sec:related_work}

\subsection{Urban Scientific Discovery}
\label{subsec:urban_discovery}

Understanding how urban form influences outcomes such as safety, equity, and accessibility is central to transportation and urban science~\cite{acuto2018building,hall2012handbook,ma2018estimating}. Traditional approaches rely on domain experts to define variables and construct statistical models that link built environment features to societal outcomes~\cite{santamouris2013energy,ma2018statistical}. For example, studies in road safety have examined how infrastructure elements, such as pedestrian crossings, bike lanes, or traffic calming measures, relate to crash risk~\cite{yu2024can,ewing2009built}.

However, these conventional workflows face significant hurdles in uncovering novel, interpretable insights from the complex urban milieu. A primary challenge is the manual and intuition-driven nature of hypothesis generation, which is often slow, susceptible to researchers' confirmation biases, and may overlook unconventional relationships~\cite{xia2025reimagining}. This reliance on pre-existing knowledge or limited observations can constrain the breadth of scientific inquiry.
Furthermore, while advanced machine learning, particularly deep learning, has shown promise in predictive tasks using urban data, such models often function as ``black boxes''~\cite{goodfellow2016deep}. Their internal representations are typically opaque, making it difficult to understand which specific factors drive predictions or to extract actionable, causal insights for urban planning and policy-making. This lack of transparency can hinder trust and adoption, especially in high-stakes decisions~\cite{benara2024crafting}.

Another critical limitation is the underutilization of rich, unstructured data sources. SVIs encapsulate vast amounts of visual information about the urban environment, from infrastructure quality to perceived safety cues~\cite{biljecki2021street}. Yet, their integration into quantitative analysis is hampered by challenges in image acquisition consistency, quality, spatial-temporal variability, and the difficulty of systematically extracting meaningful, structured variables~\cite{tang2024itinera}. Existing efforts to automate feature extraction from SVIs often rely on standard computer vision models that may not capture the nuanced, context-specific attributes relevant to complex societal outcomes without significant task-specific fine-tuning or annotation.

Recent explorations into AI-driven scientific discovery have begun to address some of these issues. For example, frameworks are emerging that use large language models for causal inference in urban contexts~\cite{xia2025reimagining} or to assist in generating hypotheses in other scientific fields by leveraging knowledge graphs alongside LLMs~\cite{lopez2025enhancing}. These works highlight a growing recognition of AI's potential to augment the discovery process, yet a dedicated, interpretable framework for hypothesis inference directly from SVIs remains an open area. Our work seeks to bridge this gap by leveraging MLLMs to systematically generate and test interpretable hypotheses about the urban environment's impact on road safety.

\subsection{Multimodal Large Language Models}
\label{subsec:mllms}

The advent of Large Language Models (LLMs) has significantly advanced capabilities in natural language understanding, generation, and reasoning~\cite{achiam2023gpt,yue2025does,wei2022chain}. Building upon this foundation, Multimodal Large Language Models (MLLMs) have emerged, extending these powerful reasoning abilities to encompass multiple modalities, most notably vision and language~\cite{zhang2024mm,yang2023dawn}. These models are designed to jointly process and interpret information from textual descriptions and visual inputs, such as images or videos.

Typical MLLM architectures integrate a pre-trained vision encoder with a pre-trained LLM~\cite{wang2025parameter}. A crucial component is the vision-language connector module, which projects visual features into a space compatible with the LLM's word embeddings. This connector can range from a simple linear projection layer, as seen in early models like LLaVA~\cite{liu2023visual}, which can perform instruction-aware visual feature extraction targeted by textual queries~\cite{kuang2025natural}. Training MLLMs often involves a multi-stage process: an initial pre-training phase to align visual and language representations using image-text datasets, followed by fine-tuning on multimodal instruction-following datasets to enhance their ability to perform specific tasks and engage in dialogue~\cite{radford2021learning}.

MLLMs have demonstrated remarkable capabilities across a wide range of tasks, including visual question answering (VQA), image captioning, multimodal dialogue, and complex visual reasoning~\cite{antol2015vqa,thawakar2025llamav}. They can generate nuanced textual descriptions of images, answer questions about visual content, and follow instructions that require grounding language in visual information. This ability to extract semantically meaningful concepts from visual scenes and align them with natural language is central to their potential. Recent research also explores techniques like Optimal Transport to achieve more interpretable semantic alignment between modalities, allowing for insights into the MLLM's reasoning process by visualizing how visual and textual elements correspond~\cite{huh2024platonic}. This is particularly relevant for applications requiring trustworthiness.

The application of MLLMs to automated scientific discovery is a burgeoning field, with studies exploring their use for generating novel research ideas, designing experiments, and even assisting in writing scientific papers~\cite{gottweis2025towards}. In urban contexts, vision-language models have been used for tasks such as function inference from street-level imagery~\cite{huang2024zero}. However, the dominant focus in automated scientific discovery has often been on the novelty or efficiency of hypothesis generation, rather than on the interpretability of the generated hypotheses or the variables used, especially when derived from complex visual data in specific domains like urban science. Our framework distinctively proposes using MLLMs not just as predictors or general-purpose reasoners, but as semantic engines to derive interpretable, human-aligned variables directly from unstructured visual data (SVIs) for statistically rigorous hypothesis inference concerning road safety.

\begin{figure*}[t]
    \centering
    \includegraphics[width=\linewidth]{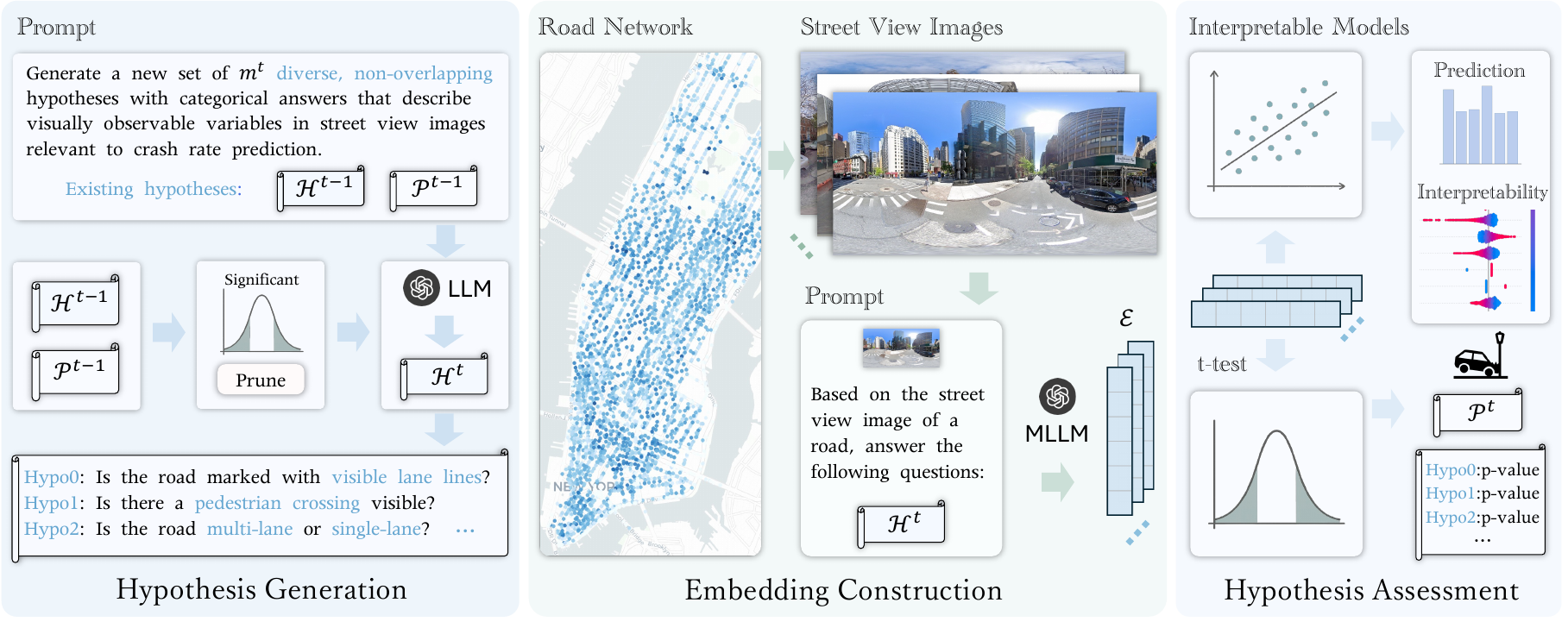}
    \caption{The \model framework consists of three iterative modules: (1) Hypothesis Generation using LLMs, (2) Embedding Construction via MLLM-based VQA on SVIs, and (3) Hypothesis Assessment using interpretable regression analysis.}
    \label{fig:framework}
\end{figure*}

\section{Methodology}
Our proposed framework, \model, provides a general approach for interpretable, hypothesis-driven scientific discovery that integrates both structured and unstructured urban data. While the framework is broadly applicable to various domains of urban mobility and societal outcomes, in this study, we focus on a specific instance: discovering road safety factors using street view imagery (SVI). Specifically, we apply \model to the problem of modeling segment-level crash risk in Manhattan, using only unstructured visual inputs as the source of explanatory information. Rather than relying on predefined visual features, \model formulates the task as inference over a space of natural language hypotheses, each representing a semantically meaningful claim about the urban environment. These hypotheses are operationalized through MLLMs and evaluated using classical statistical tools, enabling transparent modeling of transportation outcomes. We next describe the formal structure of this approach.

\subsection{Overview}

Let $\mathcal{D} = \{(x_i, y_i)\}_{i=1}^n$ denote a dataset of $n$ SVIs $x_i$ and their associated road-level crash rates $y_i \in \mathbb{R}$. 
We define a hypothesis space $\mathcal{H}$ comprising all natural-language queries that describe visually observable variables potentially related to road safety. 
Our objective is to uncover an optimal subset of hypotheses $\mathcal{H}^* = \{h_1, h_2, \dots, h_k\} \subset \mathcal{H}$ that captures meaningful visual observable features from each SVI and enables interpretable, accurate prediction of $y_i$.
We formalize this as a posterior mode estimation problem over the hypothesis space:
\begin{equation}
\mathcal{H}^* = \arg\max_{\mathcal{H}' \subseteq \mathcal{H}} P(\mathcal{H}' \mid \mathcal{D}) \propto P(\mathcal{D} \mid \mathcal{H}') \cdot P(\mathcal{H}'),
\end{equation}
where $\mathcal{H}'$ is a candidate hypothesis subset. The likelihood $P(\mathcal{D} \mid \mathcal{H}')$ captures how well the hypothesis-derived variables explain variation in crash rates, typically assessed via a regression model. The prior $P(\mathcal{H}')$ encodes structural preferences over hypothesis subsets and is implicitly governed by the generative behavior of the MLLM.
Each hypothesis $h_j \in \mathcal{H}^*$ corresponds to a semantically meaningful question with a categorical answer that could be inferred from an SVI using an MLLM. Applying these $k$ hypotheses to each image $x_i$ yields a $k$-dimensional interpretable embedding $\phi(x_i, \mathcal{H}^*) \in \mathbb{R}^k$, where each component reflects the MLLM’s answer to the corresponding hypothesis. We denote the complete embedding matrix as $\mathcal{E} \in \mathbb{R}^{n \times k}$, where $e_i = \phi(x_i, \mathcal{H}^*)$ is the embedding vector for the $i$-th image.

Bayesian inference over all possible hypothesis subsets is computationally infeasible due to the combinatorial size of $\mathcal{H}$ and the lack of a tractable likelihood model. Instead, we adopt an approximate inference strategy and frame the task as a nonparametric structure approximation problem. Starting from an initial hypothesis set $\mathcal{H}^0$ sampled from an LLM, we iteratively refine the set by evaluating each hypothesis using a linear regression model. For each hypothesis $h_j$, we assess the statistical significance of its corresponding regression coefficient via a two-sided $t$-test under the null hypothesis that the coefficient equals zero. This yields a $p$-value vector $\mathcal{P} = \{p_1, p_2, \dots, p_k\}$, where each $p_j$ quantifies the probability of observing the estimated coefficient under the null. Hypotheses with $p_j > \alpha$ (typically $0.05$) are considered statistically insignificant and are discarded. 
New hypotheses are generated to replace those that lack statistical support, with the generation process conditioned on previous significance patterns. This iterative refinement leverages empirical evidence to guide hypothesis proposals, approximating posterior inference over the hypothesis space.
The overall process forms an iterative loop of hypothesis generation, assessment, and refinement, guided by both LLM priors and statistical evidence. An overview of the \model framework is illustrated in Figure~\ref{fig:framework}.

\subsection{Hypothesis Generation}

A fundamental challenge in urban and transportation research is formulating testable hypotheses that link observable characteristics of the built environment to outcomes such as safety, mobility, or equity. Traditional approaches often depend on domain experts to define variables or pose questions, which limits scalability, introduces subjectivity, and constrains discovery. To address these limitations, we leverage LLMs as generative engines for proposing semantically rich, visually grounded hypotheses expressed in natural language. This enables a more systematic and scalable approach to variable discovery from unstructured urban data sources such as SVIs.

At each iteration $t$, the framework refines the hypothesis set $\mathcal{H}^{t-1}$ using statistical evidence derived from the previous assessment. For each hypothesis $h_j \in \mathcal{H}^{t-1}$, we compute a $p$-value $p_j$ using a two-sided $t$-test on the coefficient estimated by a regression model, where the input variable is derived from the MLLM-inferred categorical responses to $h_j$ across all SVIs. The detailed procedure for constructing hypothesis-driven embeddings is described in a later subsection. Hypotheses with $p_j > \alpha$ (typically $\alpha = 0.05$) are considered statistically insignificant. While the prompt for the LLM includes the full set of previous hypotheses $\mathcal{H}^{t-1}$ and their $p$-values $\mathcal{P}^{t-1}$, only $m^t$ new hypotheses are generated, where $m^t$ equals the number of pruned hypotheses. This maintains a fixed hypothesis set size while ensuring that each iteration incorporates empirical feedback into the generative process. Formally, the hypothesis generation step is given by:
\begin{equation}
\mathcal{H}^{t} \sim \mathrm{LLM}\left(\mathrm{Prompt}_{\textit{HypoGen}}(\mathcal{H}^{t-1}, \mathcal{P}^{t-1}, m^t)\right),
\end{equation}
where $\mathrm{Prompt}_{\textit{HypoGen}}$ encodes the task instructions, previous hypothesis outcomes, and domain constraints. The updated hypothesis set is formed by retaining significant hypotheses from $\mathcal{H}^{t-1}$ and appending the newly generated ones.

To ensure that hypothesis generation is both semantically controlled and statistically informed, we design a unified prompt that combines task instructions, contextual input, and generation constraints. The prompt defines the hypothesis generation task in the context of segment-level crash rate, instructing the model to produce categorical, visually inferable questions grounded in SVIs. It includes the full set of hypotheses from the previous iteration along with their $p$-values, enabling the LLM to learn from both statistically significant and insignificant examples. Finally, it specifies clear design rules to enforce uniqueness, clarity, and visual observability. A conceptual version of the prompt is shown below:
\vspace{5pt}
\begin{tcolorbox}[title=Conceptual Prompt for Hypothesis Generation, breakable]
\begin{minted}{markdown}
You are a transportation researcher studying the relationship between visual urban features and crash risk. Your task is to generate a new set of interpretable, visually grounded hypotheses in the form of questions with categorical answers. Each hypothesis should describe a feature that can be inferred directly from street view imagery and that could help explain variation in segment-level crash rates.

### Existing hypotheses:  
```{hypo t-1}```

### Design guidelines:
1. Each hypothesis should be unique and avoid semantic overlap with the existing ones.
2. Hypotheses must be answerable using only the visual content of a street view image, without relying on metadata or external context.
3. Each hypothesis should correspond to an observable environmental feature that could plausibly influence road safety.
4. Formulate each hypothesis with a small set of mutually exclusive and collectively exhaustive categorical answer choices.
5. Avoid vague, subjective, or ambiguous terms. Focus on concrete visual concepts such as road geometry, traffic controls, pedestrian infrastructure, visibility conditions, or physical obstructions.

### Your response should include:
1. A short reasoning paragraph reflecting on the strengths and gaps of the existing hypotheses and proposing new directions for expansion.
2. A list of newly proposed hypotheses, each followed by its categorical answer options in dictionary format.
\end{minted}
\end{tcolorbox}
\vspace{5pt}
This prompt design leverages the LLM’s world knowledge to formulate hypotheses that are both visually grounded and relevant to road safety. By conditioning on prior hypotheses and their statistical outcomes, the model is encouraged to generate semantically diverse and empirically plausible questions that can be reliably inferred from SVIs.
The complete implementation of the prompt is provided in Appendix~\ref{appendix:prompt}. 
In addition to the conceptual version shown above, the implemented prompt includes additional instructions that control the output format and enforce structural consistency. These enhancements are designed to improve the stability and reproducibility of the LLM’s outputs across iterations, ensuring a reliable and interpretable hypothesis generation process.

This iterative, LLM-in-the-loop design naturally aligns with the principles of Approximate Bayesian Computation (ABC)~\cite{csillery2010approximate,sunnaaker2013approximate}. In this view, the LLM acts as a nonparametric simulator over the space of natural-language hypotheses. Instead of explicitly specifying or sampling from a tractable likelihood function, the system leverages the LLM to propose candidate hypotheses conditioned on previously observed data and their corresponding statistical evaluations. The inclusion of both significant and non-significant hypotheses, along with their $p$-values, enables the LLM to implicitly learn which semantic structures tend to yield stronger associations with crash outcomes. The statistical model then plays the role of an empirical evaluator, determining which hypotheses are retained based on significance thresholds. This iterative process approximates posterior inference by refining the hypothesis set through data-informed generation and evidence-based pruning. This structure enables interpretable, statistically grounded exploration of complex visual environments and supports the discovery of novel variables that are relevant for transportation planning, safety analysis, and urban policy design.

\subsection{Embedding Construction}

To enable hypothesis-driven modeling, the set of generated hypotheses $\mathcal{H}^t = \{h_1^t, h_2^t, \dots, h_k^t\}$ must be transformed into structured representations that can serve as explanatory variables in statistical models. While conventional vision-based approaches typically extract latent features from images using deep neural networks, such representations are often opaque and difficult to interpret. In contrast, our framework constructs an interpretable embedding space that is explicitly aligned with the semantics of the generated hypotheses.
For each SVI $x_i$, we employ an MLLM to infer categorical responses to each hypothesis $h_j^t \in \mathcal{H}^t$, based solely on the visual content of the SVI. Each response corresponds to a discrete category or other mutually exclusive options defined by the hypothesis. These responses are then encoded into a $k$-dimensional vector $e_i^t \in \mathbb{R}^k$, where the $j$-th element represents the MLLM-inferred answer to hypothesis $h_j^t$ for image $x_i$.
Formally, we define the embedding process as:
\begin{equation}
    e_i^t \sim \mathrm{MLLM}\left(x_i, \mathrm{Prompt}_{\textit{Embed}}(\mathcal{H}^t)\right),
\end{equation}
where $\mathrm{Prompt}_{\textit{Embed}}(\mathcal{H}^t)$ encodes the set of hypotheses into a format suitable for batched MLLM inference. This prompt instructs the model to generate structured categorical answers for all hypotheses in $\mathcal{H}^t$ conditioned on the visual content of $x_i$. The resulting embeddings retain full semantic traceability to the original hypotheses, allowing each variable to be interpreted as the operationalization of a natural-language question about the SVI.

The complete set of embeddings $\{e_1^t, \dots, e_n^t\}$ defines a structured matrix $\mathcal{E}^t \in \mathbb{R}^{n \times k}$ that serves as the input to downstream statistical analysis. Unlike latent visual features, each column in $\mathcal{E}^t$ corresponds to a semantically meaningful hypothesis, thereby enabling interpretable regression modeling, significance testing, and iterative refinement. This embedding mechanism forms a core component of the framework’s transparency and supports the goal of integrating unstructured visual data into statistically grounded urban analysis. An overview of this process is depicted in the center panel of Figure~\ref{fig:framework}.

\subsection{Hypothesis Assessment}

Given the hypothesis-aligned embedding matrix $\mathcal{E}^t \in \mathbb{R}^{n \times k}$ constructed from the current hypothesis set $\mathcal{H}^t$, the next step is to evaluate the empirical relevance of each hypothesis in explaining variation in segment-level crash rates. Rather than optimizing for predictive performance alone, our objective is to support transparent, interpretable modeling that enables attribution of outcomes to individual hypotheses. This is particularly important in transportation and urban policy contexts, where analytical traceability and explanatory clarity are essential for decision-making and public accountability.
To this end, we adopt linear regression as the primary modeling framework. Each column of $\mathcal{E}^t$ corresponds to a hypothesis-specific variable derived from categorical responses generated by the MLLM. We fit a linear model of the form:
\begin{equation}
    y_i = \beta_0 + \sum_{j=1}^{k} \beta_j e_{ij}^t + \varepsilon_i,
\end{equation}
where $y_i$ is the observed crash rate for SVI $x_i$, $e_{ij}^t$ is the value of the $j$-th embedding dimension for the $i$-th SVI, $\beta_j$ is the corresponding regression coefficient, and $\varepsilon_i$ is an independent error term assumed to be normally distributed.

We then apply a two-sided $t$-test to each coefficient $\beta_j$ to assess the null hypothesis that $\beta_j = 0$, using standard errors estimated from the fitted model. This yields a $p$-value $p_j^t$ that quantifies the likelihood that the observed effect could arise under the null. Collectively, the vector $\mathcal{P}^t = \{p_1^t, p_2^t, \dots, p_k^t\}$ summarizes the statistical significance of each hypothesis in $\mathcal{H}^t$. Hypotheses with $p_j^t > \alpha$ (typically $\alpha = 0.05$) are considered statistically insignificant and are pruned in the next round. The number of such hypotheses, $m^t$, determines how many new hypotheses are generated in the subsequent iteration. Fitting the regression model $\mathcal{M}$ to the embedding matrix $\mathcal{E}^t$ yields both fitted outcomes and statistical estimates of hypothesis relevance:
\begin{equation}
    \{\widehat{y}_i\}_{i=1}^{n},\ \{\widehat{\beta}_j,\, p^t_j\}_{j=1}^{k} \gets \mathcal{M}(\mathcal{E}^t),
\end{equation}
where $\mathcal{M}$ denotes the linear regression model applied to the embedding matrix $\mathcal{E}^t$.

This assessment step plays two complementary roles within the overall framework. First, it provides a quantitative basis for interpreting the influence of each hypothesis on crash outcomes, enabling transparent attribution and variable importance comparisons. Second, it functions as a mechanism for iterative hypothesis refinement, systematically pruning low-utility hypotheses and informing the next round of LLM-based generation. The right panel of Figure~\ref{fig:framework} illustrates this process within the broader iterative pipeline.

\subsection{Iterative Posterior Approximation}

\begin{algorithm}[h]
\caption{Iterative Posterior Approximation}
\label{alg:iterative}
\begin{algorithmic}[1]
\Require Dataset $\mathcal{D} = \{(x_i, y_i)\}_{i=1}^n$; number of total hypotheses $k$; number of iterations $T$; interpretable model $\mathcal{M}$
\Ensure Final hypothesis set $\mathcal{H}^T$ and embedding matrix $\mathcal{E}^T$
\State Initialize $\mathcal{H}^0 \sim \mathrm{LLM}(\mathrm{Prompt}_{\textit{HypoGen}}(k))$
\For{$t = 1, 2, \dots, T$}
    \State $\mathcal{H}^t \sim \mathrm{LLM}(\mathrm{Prompt}_{\textit{HypoGen}}(\mathcal{H}^{t-1}, \mathcal{P}^{t-1}, m^t))$ \hfill $\triangleright $\texttt{ Hypothesis Generation}
    \State $\mathcal{E}^t = \left\{e_i^t = \mathrm{MLLM}(x_i, \mathrm{Prompt}_{\textit{Embed}}(\mathcal{H}^t)) \right\}_{i=1}^n$ \hfill $\triangleright $\texttt{ Embedding Construction}
    \State $\{\widehat{y_i}\}_{i=1}^{n}, \mathcal{P}^t \gets \mathcal{M}(\mathcal{E}^t)$ \hfill $\triangleright $\texttt{ Hypothesis Assessment}
\EndFor
\end{algorithmic}
\end{algorithm}

The proposed framework operates through an iterative loop that alternates between hypothesis generation, embedding construction, and hypothesis assessment. This process approximates posterior inference over a combinatorially large and semantically structured hypothesis space, where each hypothesis represents a visually observable attribute inferred from SVIs and encoded as an interpretable variable.

At each iteration, the goal is to refine the hypothesis set $\mathcal{H}^t$ such that it better explains or predicts the target outcome. Although this optimization problem is inherently nonconvex and lacks a tractable likelihood, our framework approximates it through data-driven pruning and targeted generation. By leveraging empirical signals to guide each update, the hypothesis set progressively improves in relevance and predictive utility. Additionally,
unlike conventional optimization techniques such as gradient descent~\cite{ruder2016overview} or expectation maximization~\cite{moon1996expectation}, which operate under well-defined objective functions and convergence guarantees, our setting relies on sampling from a nonparametric, LLM-driven hypothesis space. This introduces uncertainty in the refinement trajectory. To address this, we adopt a conservative update strategy: newly proposed hypotheses are retained only if they improve predictive performance on a validation set relative to the previous iteration.

The full procedure is summarized in Algorithm~\ref{alg:iterative}. In each iteration, statistically insignificant hypotheses from $\mathcal{H}^{t-1}$ are identified based on their $p$-values $\mathcal{P}^{t-1}$ and pruned. The remaining hypotheses provide contextual input for LLM-based generation of $m^t$ new candidates. These are used to construct a hypothesis-aligned embedding matrix $\mathcal{E}^t$ via MLLM reasoning, which is then passed to a linear model for regression. This loop continues for a fixed number of iterations or until the convergence criteria are met.

This iterative posterior approximation framework provides a unified mechanism for discovering semantically meaningful, statistically validated variables from unstructured data. By embedding hypothesis generation within a statistically informed feedback loop, \model enables a scalable alternative to manual variable engineering, particularly suited to domains like transportation safety, where interpretability, traceability, and data diversity are paramount. While our implementation focuses on segment-level crash risk and SVIs, the underlying framework is general and can be adapted to other urban outcomes and multimodal data settings.

Although \model does not compute an explicit posterior distribution, it draws conceptual inspiration from Approximate Bayesian Computation (ABC). By combining simulation through LLM-based hypothesis generation and empirical evaluation via regression-based significance testing, the framework approximates posterior inference over hypothesis structures without relying on a tractable likelihood. This enables a principled yet flexible approach to variable discovery in high-dimensional, semantically rich domains.

\section{Experiments}
\label{sec:exp}

\subsection{Settings}

To evaluate the effectiveness of \model in supporting interpretable and data-driven discovery of urban road safety factors, we conduct experiments on road segments within Manhattan, New York City. This area provides a complex urban setting characterized by high traffic density, varied land use, and extensive open data resources. Our objective is to predict and explain segment-level crash risk based solely on visual inputs from street-view SVI, without the use of predefined variables or manual feature annotation.

Crash risk is quantified using a standardized crash rate~\cite{hou2020correlated,zeng2017multivariate,yu2024can}, defined for each segment as:
\begin{equation}
CR_i = \frac{\text{No\_crash}_i}{AADT_i \times L_i \times \frac{365}{1,000,000}},
\end{equation}
where $\text{No\_crash}_i$ denotes the annual average number of reported crashes, $AADT_i$ represents the average annual daily traffic volume, and $L_i$ is the length of the road segment in kilometers. This formulation adjusts for traffic exposure and segment size, and is widely adopted in transportation safety research to enable fair comparisons across road types and traffic conditions.

Crash records were sourced from NYC Open Data\footnote{\url{https://opendata.cityofnewyork.us/}}, and traffic volume data ($AADT$) was obtained from the New York State Department of Transportation\footnote{\url{https://dos.ny.gov/location/new-york-state-department-transportation}}. Street-view imagery was collected using ArcGIS~\cite{wong2005statistical}, with panoramic images sampled every 15 meters along road centerlines and retrieved via the Google Street View API\footnote{\url{https://developers.google.com/maps/documentation/streetview/overview}}. After preprocessing and filtering, the dataset includes 16,000 images for training, 2,000 for validation, and 2,000 for testing.

For hypothesis generation, we use GPT-4o~\cite{hurst2024gpt}\footnote{\url{https://platform.openai.com/docs/models}}. To construct visual embeddings based on structured prompts, we use InternVL2.5-78B~\cite{chen2024expanding}\footnote{\url{https://huggingface.co/OpenGVLab/InternVL2_5-78B}}. All multimodal models are deployed using LMDeploy~\cite{2023lmdeploy}, which provides an efficient and reproducible framework for serving large-scale MLLMs.

To provide a comparative reference for evaluating our approach, we also compile a comprehensive set of 58 built environment features, drawn from five domains: (1) road design (e.g., width, highway indicator), (2) land use composition and entropy, (3) point-of-interest (POI) features, (4) traffic-related facilities (e.g., crossings, bus stops), and (5) visual indices derived from panoptic segmentation (e.g., vegetation, building, road proportions). These variables serve as a baseline representation of conventional urban form and are used in a post hoc analysis to assess the added value of the discovered hypotheses. A complete list of variables and data sources is provided in Appendix~\ref{appx:bvar}.

In our primary modeling pipeline, however, these handcrafted features are not included. Instead, we rely exclusively on hypotheses generated by the language model and their corresponding embeddings inferred by the MLLMs. This design ensures that predictive insights emerge solely from automatically discovered, visually interpretable patterns, allowing us to assess the capacity of \model to support transparent, scalable, and data-driven scientific discovery grounded in the visual environment.

\begin{figure}[t]
    \centering
    \includegraphics[width=0.6\linewidth]{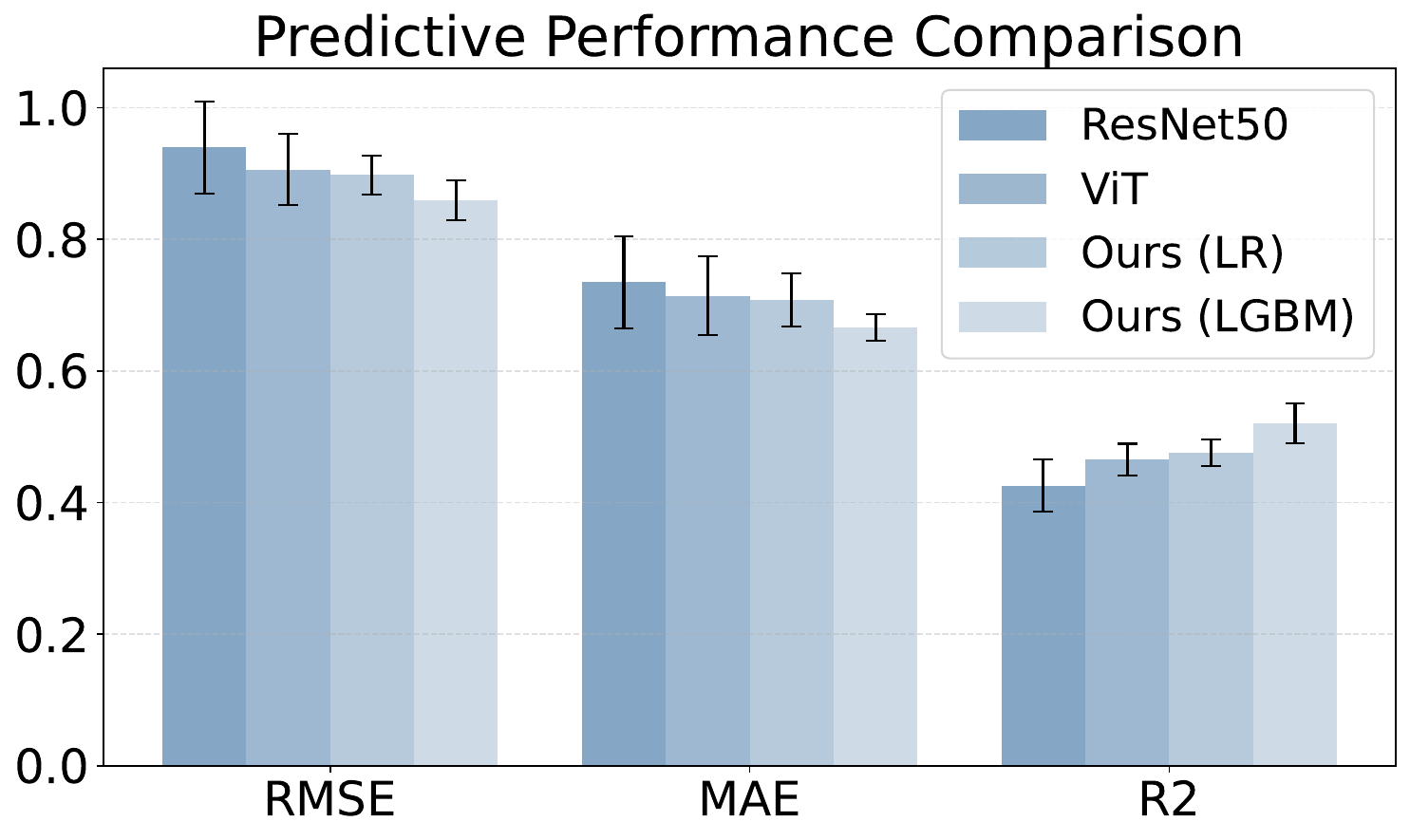}
    \caption{Performance comparison between ResNet, ViT, and our interpretable embedding-based models using linear regression (LR) and LightGBM (LGBM). Lower MAE ($\downarrow$) and RMSE ($\downarrow$), and higher $R^2$ ($\uparrow$), indicate better performance. Our method achieves the best results across all three metrics.}
    \label{fig:performance_comparison}
\end{figure}

\subsection{Predictive Performance}

We begin by evaluating the ability of our interpretable embedding framework to predict segment-level crash rates using only visual information extracted from SVIs. Figure~\ref{fig:performance_comparison} compares our approach with standard vision-based baselines across three widely used metrics: root mean squared error (RMSE), mean absolute error (MAE), and coefficient of determination ($R^2$). These metrics are formally defined as follows:
\begin{align}
\text{RMSE} &= \sqrt{\frac{1}{n} \sum_{i=1}^n (y_i - \widehat{y}_i)^2}, \\
\text{MAE}  &= \frac{1}{n} \sum_{i=1}^n |y_i - \widehat{y}_i|, \\
R^2 &= 1 - \frac{\sum_{i=1}^n (y_i - \widehat{y}_i)^2}{\sum_{i=1}^n (y_i - \bar{y})^2},
\end{align}
where $y_i$ is the observed crash rate for the $i$-th SVI, $\widehat{y}_i$ is the corresponding predicted value, $\bar{y}$ is the mean of all observed values, and $n$ is the total number of SVIs.

For the baselines, we fine-tune two representative image encoders to directly regress crash rates from raw images: ResNet-50, a widely adopted convolutional neural network, and ViT-B/16, a transformer-based architecture that segments each image into $16 \times 16$ patches for self-attention processing. These models represent the current standard in vision-based modeling for urban applications.

Our proposed framework replaces latent visual features with structured, interpretable embeddings derived from MLLM responses to hypothesis-driven prompts. We evaluate two variants: one using linear regression (LR)~\cite{montgomery2021introduction} and another using LightGBM (LGBM)~\cite{ke2017lightgbm} as the downstream predictor. Both variants rely on embeddings aligned with semantically meaningful hypotheses about the built environment, ensuring transparency and domain relevance.

\begin{figure}[t]
    \centering
    \includegraphics[width=.75\linewidth]{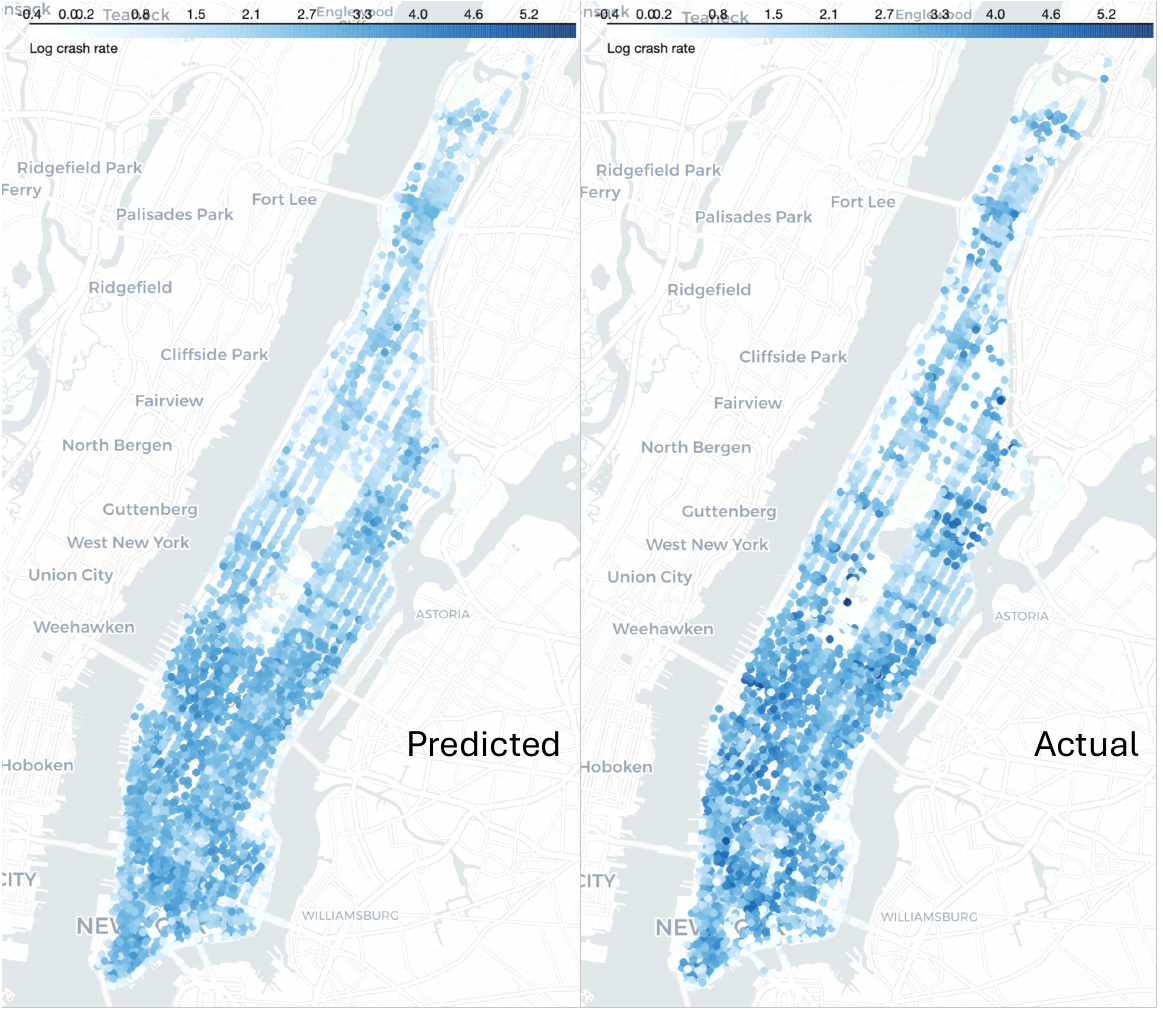}
    \caption{Spatial distribution of predicted (left) vs. actual (right) crash rates (log scale) across Manhattan road segments. \model captures key spatial patterns in crash risk, including high-risk areas in lower and midtown Manhattan.}
    \label{fig:map_comparison}
\end{figure}

As shown in Figure~\ref{fig:performance_comparison}, our approach achieves superior performance across all evaluation metrics. Notably, the LGBM-based variant yields the lowest prediction error and highest explained variance, outperforming both deep learning baselines despite using only interpretable, categorical features. This result highlights the effectiveness of hypothesis-driven embeddings in capturing safety-relevant visual cues and supports the viability of interpretable modeling for transportation safety analysis. By enabling both statistical rigor and semantic traceability, our method contributes to more accountable and policy-relevant crash risk modeling in urban contexts.

To examine whether the model generalises beyond pointwise accuracy and reproduces the spatial structure of crash risk, Figure~\ref{fig:map_comparison} presents a comparison between predicted and actual log crash rates at the road segment level across Manhattan. The predictions are produced using five-fold cross-validation to ensure out-of-sample generalization, where each road segment is predicted only when held out from training. Visually, the predicted crash risk surface exhibits a high degree of spatial coherence with the ground truth. Both maps capture broad urban gradients, with elevated crash rates concentrated in denser southern and midtown zones. While some local differences remain, especially at the segment level, the model appears to reproduce neighborhood-level trends and spatial clusters with reasonable fidelity. This supports the framework’s capacity to generalize across heterogeneous street environments while preserving interpretable spatial structure relevant for transportation planning and safety policy.

\subsection{Discovered Factors}
\label{exp:df}

\begin{figure}[h]
    \centering
    \includegraphics[width=.9\linewidth]{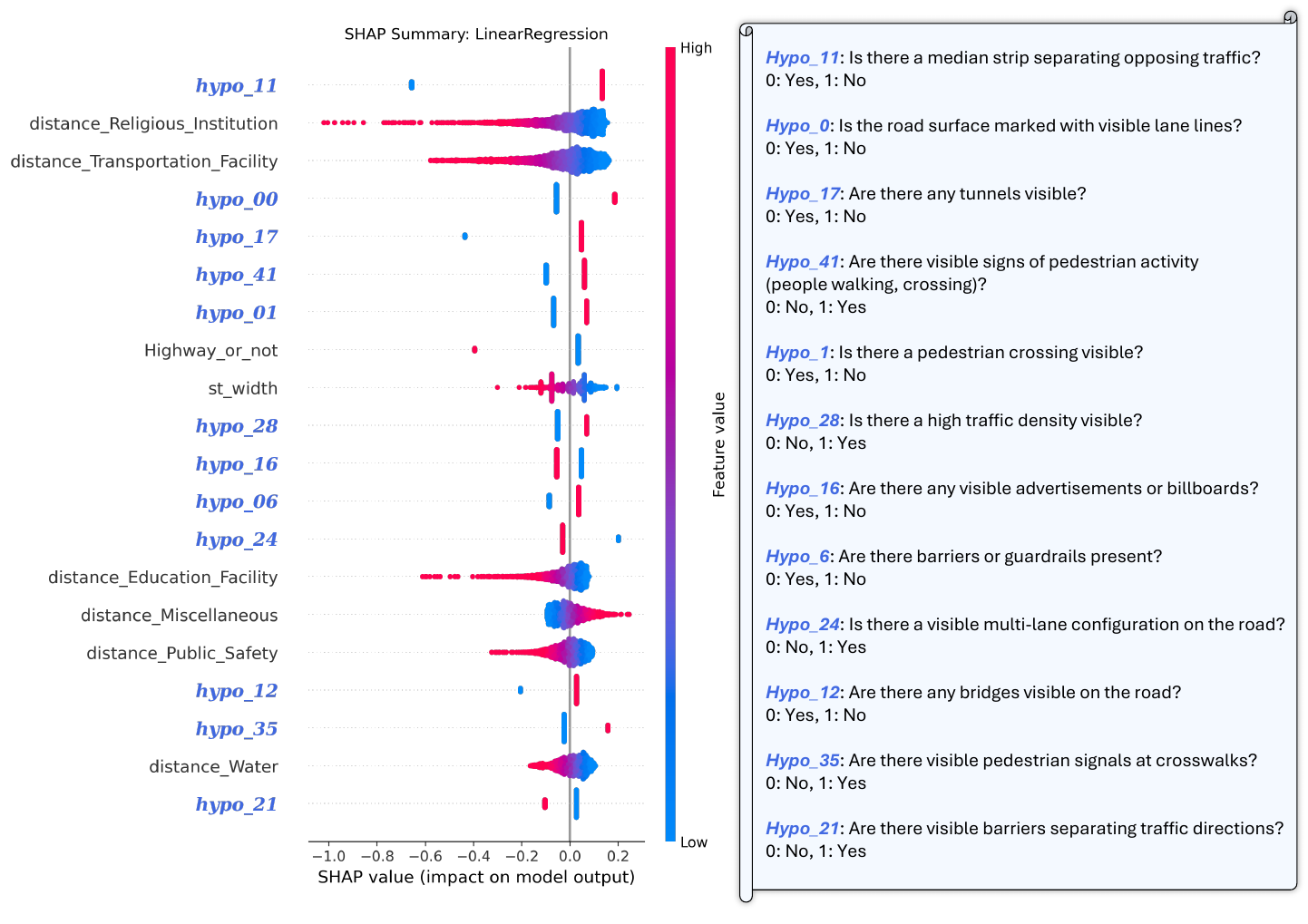}
    \caption{SHAP summary plot of the regression model with both traditional built environment variables and discovered hypotheses. The right panel maps the top hypothesis variables to their natural-language question meanings.}
    \label{fig:shap_summary}
\end{figure}

A core objective of \model is to move beyond predictive accuracy and toward interpretable, data-driven discovery of visual factors that are often missing from conventional urban analytics. To evaluate the substantive relevance of these learned variables, we apply SHAP (SHapley Additive exPlanations)~\cite{lundberg2017unified} to a regression model trained on both traditional built environment features and hypothesis-derived embeddings. This allows us to quantify the marginal contribution of each variable to the prediction of segment-level crash rates.

Figure~\ref{fig:shap_summary} displays the top 20 features ranked by mean absolute SHAP value. These include both conventional indicators and hypotheses discovered by the LLM-based generation process. Notably, the majority of the most influential features are derived from \model’s interpretable embedding pipeline. This underscores the framework’s capacity to uncover significant, human-interpretable factors that complement or extend beyond standard structured datasets, reinforcing its potential as a tool for empirical urban discovery rather than a purely predictive black box.

Many of the hypotheses uncovered by \model are consistent with well-established principles in transportation safety research, demonstrating the model’s ability to recover latent domain knowledge without manual encoding. For example, \texttt{Hypo\_11} (“Is there a median strip separating opposing traffic?”) and \texttt{Hypo\_0} (“Is the road surface marked with visible lane lines?”) are both among the most influential variables, with negative SHAP contributions when absent. Their presence is associated with reduced crash risk, aligning with conventional traffic engineering insights regarding lane separation and visual guidance.

Beyond these expected patterns, \model also identifies finer-grained or underrepresented elements related to pedestrian safety. Hypotheses such as \texttt{Hypo\_1} (crosswalk presence), \texttt{Hypo\_41} (pedestrians visible), and \texttt{Hypo\_35} (pedestrian signals present) reflect aspects of pedestrian infrastructure and activity that are context-dependent and difficult to capture through existing spatial datasets. These results suggest that the hypothesis space can encode subtle, behaviorally relevant cues often missed by traditional indicators.

In addition, \model detects environmental features that are rarely operationalized in urban data systems but are visually salient. For instance, \texttt{Hypo\_16} (“Are there any advertisements or billboards?”) and \texttt{Hypo\_6} (“Are there barriers or guardrails present?”) highlight perceptual or protective elements that may influence driver attention or traffic separation but are not systematically recorded in built environment inventories.

Compared to traditional features such as segment length, land use mix, or POI features, these hypothesis-derived variables are more granular, interpretable, and directly tied to observable characteristics in the street view imagery. This illustrates the distinctive advantage of hypothesis-aligned embeddings: they provide a transparent interface between visual sensing and statistical modeling, enabling interpretable insights that are actionable for urban planning and policy. A full list of hypotheses, including semantic descriptions and statistical summaries, is provided in Appendix~\ref{appendix:hypo_analysis}.

Taken together, these findings demonstrate that \model is capable not only of achieving competitive predictive accuracy but also of surfacing novel, interpretable factors that enrich the scientific understanding of urban safety. Its ability to propose, test, and validate hypotheses from raw SVIs without manual annotation or expert-defined features, highlighting the potential of MLLM-powered frameworks to transform how we conduct research in urban and transportation science.

\subsection{Variable Significance and Independence}

\begin{figure}[t]
    \centering
    \includegraphics[width=.7\linewidth]{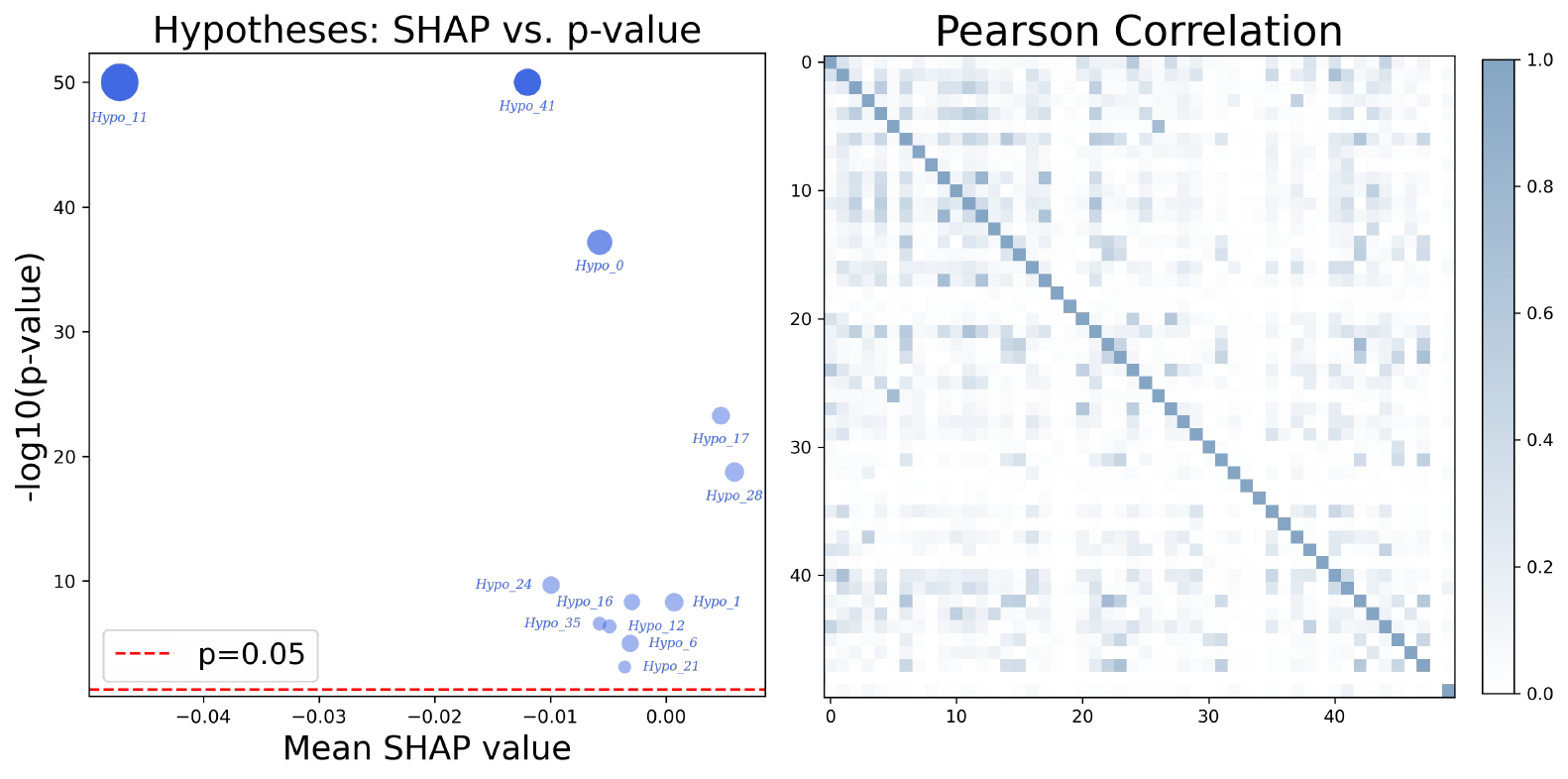}
    \caption{
    (Left) Each hypothesis is plotted by its average SHAP value and $-\log_{10}(p\text{-value})$ from regression. Variables in the top left are highly significant and predictive. (Right) Pearson correlation matrix between hypotheses, showing low pairwise correlation and structural independence.
  }
  \label{fig:shap_pval_corr}
\end{figure}

To assess the robustness and distinctiveness of the hypotheses discovered by \model, we conduct two complementary evaluations. First, we examine each hypothesis in terms of its explanatory power and statistical significance. Second, we analyze the pairwise correlation structure to evaluate redundancy and independence among the learned variables. The results of both analyses are presented in Figure~\ref{fig:shap_pval_corr}.

\paragraph{Explanatory value and statistical support.}
The left panel of Figure~\ref{fig:shap_pval_corr} visualizes each hypothesis according to its average SHAP value (x-axis), reflecting predictive contribution, and the negative base-10 logarithm of its $p$-value from linear regression (y-axis), reflecting statistical significance. This joint analysis enables a comprehensive evaluation of each hypothesis in terms of both empirical strength and inferential robustness. Hypotheses positioned in the upper-left quadrant of the plot are both highly informative and strongly supported by the data.
Among the most significant variables, \texttt{Hypo\_11}, \texttt{Hypo\_41}, and \texttt{Hypo\_0} emerge as dominant contributors. These correspond to interpretable, visually grounded questions such as whether there is a median strip separating opposing traffic, whether pedestrians are visible, and whether lane markings are present. Their prominence is consistent with established traffic safety principles and affirms the model’s ability to rediscover meaningful urban design features directly from visual data.

\paragraph{Pairwise correlation.}
The right panel of Figure~\ref{fig:shap_pval_corr} displays the Pearson correlation matrix across all hypothesis-derived variables, based on their categorical responses over the dataset. The matrix reveals minimal off-diagonal intensity, indicating generally low linear dependence between hypotheses. Quantitatively, over 85\% of all variable pairs exhibit absolute correlation coefficients below 0.2.
This low interdependence suggests that the discovered hypotheses capture complementary and non-redundant aspects of the street-level environment, including geometric layout, traffic control infrastructure, and pedestrian activity. Such structural independence enhances the reliability of downstream statistical modeling by mitigating risks associated with multicollinearity, and it strengthens the interpretability and modularity of the learned feature space, both essential for transparent urban policy analysis and evidence-based decision making.

\paragraph{Summary.}
These results show that the hypotheses generated by \model are not only individually meaningful but also collectively informative. They provide measurable explanatory value, pass statistical significance tests, and avoid redundancy. This aligns with the objectives of interpretable machine learning in transportation research and supports data-driven urban planning efforts grounded in visual understanding of street-level environments.

\subsection{Robustness and Validity}

\begin{figure}[t]
    \centering
    \includegraphics[width=.7\linewidth]{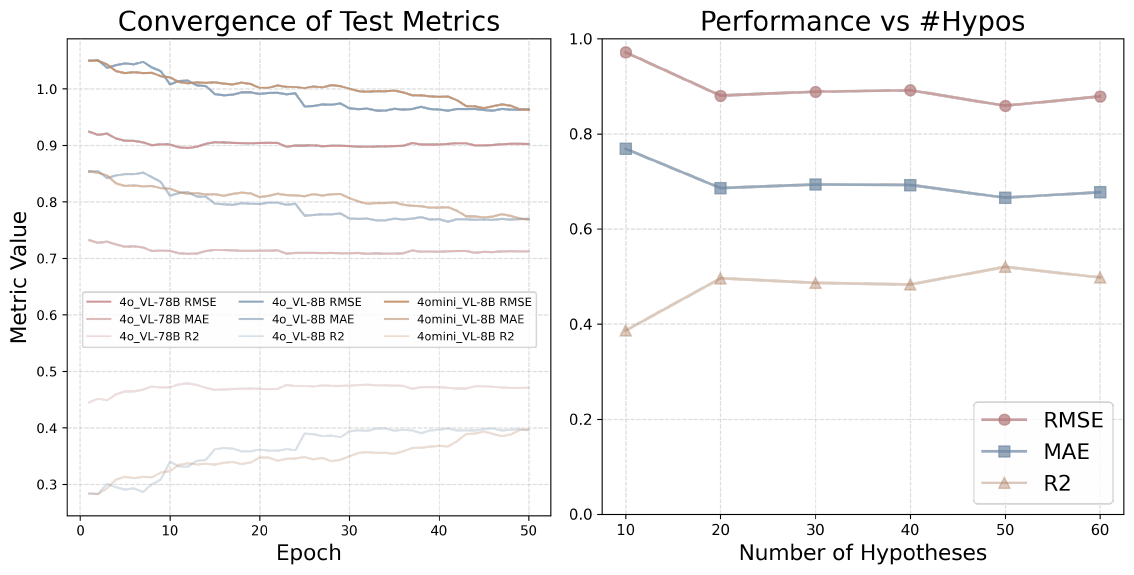}
    \caption{Robustness analysis. (Left) Convergence of test metrics across different LLM and MLLM configurations. High-capacity MLLMs (78B) yield better and faster convergence. (Right) Performance as a function of the number of hypotheses used. Optimal performance is observed at 50 hypotheses.}
    \label{fig:robustness}
\end{figure}

We assess the robustness of \model by varying three key factors: (1) the capacity of the language model (LLM) used for hypothesis generation, (2) the capacity of the vision-language model (MLLM) used for embedding construction, and (3) the number of hypotheses used to generate interpretable features. Figure~\ref{fig:robustness} presents results that illustrate how each of these choices affects convergence speed, predictive accuracy, and model stability.

\paragraph{Effect of model capacity.}
The left panel compares the convergence of test performance over 50 training epochs using different combinations of LLMs and MLLMs. Specifically, we consider GPT-4o and GPT-4o-mini as the hypothesis generators, paired with InternVL2.5 models of 8B and 78B parameters for embedding construction. Larger MLLMs (78B) consistently yield better predictive performance and faster convergence, underscoring the critical role of visual reasoning capacity in answering hypothesis queries from SVIs. For example, with InternVL2.5-78B, models converge in fewer than 10 epochs, while smaller models (8B) exhibit slower and noisier learning curves.
Language model size also affects convergence, though to a lesser extent. GPT-4o achieves faster improvements than GPT-4o-mini, but both converge to similar final performance. This suggests that for the relatively constrained task of binary hypothesis generation, smaller LLMs are sufficient, though larger models may enhance efficiency by producing more immediately useful hypotheses.

\paragraph{Effect of hypothesis-set size.}
The right panel examines performance as a function of the number of hypotheses used to construct interpretable embeddings. Performance improves steadily up to around 50 hypotheses, with diminishing returns and slight degradation beyond that point. This pattern reflects a balance between semantic expressiveness and statistical noise: too few hypotheses limit the model’s representational capacity, while too many may introduce redundancy, spurious correlations, or overfitting. The model achieves optimal RMSE, MAE, and $R^2$ values when embedding dimensionality is neither overly constrained nor saturated.

\paragraph{Implications for practice.}
These experiments demonstrate that \model is robust to reasonable changes in foundation-model capacity and to the choice of hypothesis-set size.
In resource-constrained settings, an issue frequently faced by transportation agencies, one can adopt a smaller language model without a large performance penalty, provided that a sufficiently capable vision–language model is available.
The sensitivity analysis on hypothesis cardinality also offers guidance for practitioners: start with a moderate set (40–60 items), monitor statistical significance during training, and prune or augment as needed.
Together, the results confirm that the proposed framework delivers stable, interpretable, and policy-relevant insights across a range of computational budgets and modelling choices.

\subsection{Qualitative Verification of MLLM Answers}
\label{appendix:vqa_case}

\begin{figure}[h]
    \centering
    \includegraphics[width=.8\linewidth]{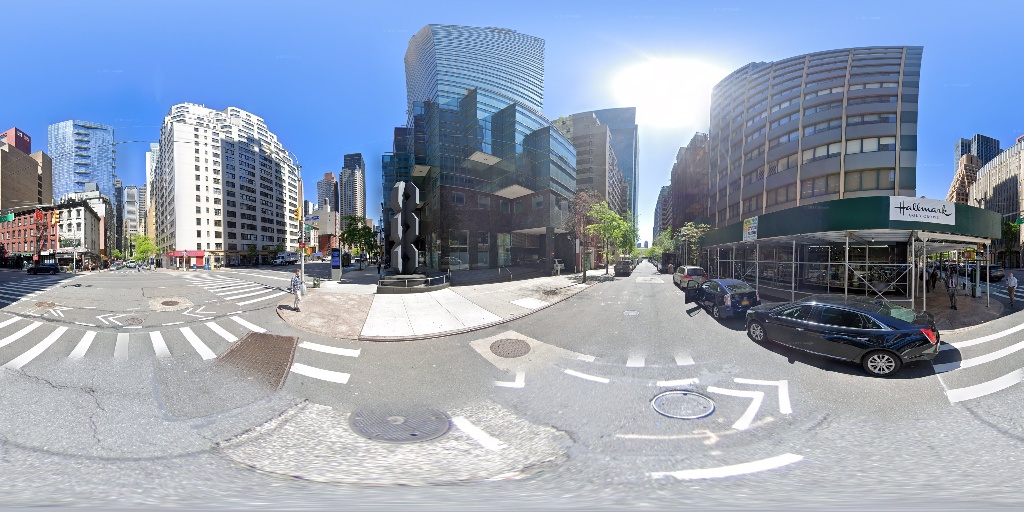}
    \caption{Panoramic SVI used for VQA analysis.}
    \label{fig:case_study_img}
\end{figure}

To evaluate the reliability of MLLM-derived answers in our framework, we conducted a manual audit on a representative panoramic SVI (Figure~\ref{fig:case_study_img}) using the final hypothesis set (see Figure~\ref{fig:full_hypo}). Each answer was cross-checked by an expert in transportation systems and labeled as \emph{correct}, \emph{partially correct}, or \emph{incorrect}. The results are summarized in Table~\ref{tab:vqa_audit}, with detailed error analysis provided in Table~\ref{tab:vqa_errors}.

\paragraph{Overall accuracy.}

As shown in Table~\ref{tab:vqa_audit}, 42 of the 50 MLLM answers (84\%) were fully correct. Three were partially correct and five were judged incorrect. Correct predictions typically involved clear, high-contrast visual elements such as lane markings, sidewalks, crosswalks, parked vehicles, and urban density, consistent with the top contributors identified by SHAP analysis in Section~\ref{exp:df}.

\begin{table}[h]
\centering
\small
\caption{Manual audit of MLLM answers for the 50 retained hypotheses.}
\label{tab:vqa_audit}
\begin{tabular}{lccc}
\toprule
\textbf{Outcome} & \textbf{\# Hypotheses} & \textbf{Examples} & \textbf{Typical cause}\\
\midrule
Correct            & 42 & lane lines, crosswalk, bus stop, median strip & clear visual cue\\
Partially correct  &  3 & vegetation obstruction, traffic density       & ambiguous threshold\\
Incorrect          &  5 & snow/ice, pedestrian signal, speed camera     & rare or small object\\
\bottomrule
\end{tabular}
\end{table}

\paragraph{Error profile.}
A breakdown of the eight non-correct answers is shown in Table~\ref{tab:vqa_errors}. Errors generally fall into three categories:

\begin{itemize}[leftmargin=*]
\item \textbf{Ambiguous semantics.} Hypotheses involving subjective thresholds (e.g., visibility obstructions or what counts as an “advertisement”) yielded borderline results due to interpretive ambiguity in both human and model judgment.
\item \textbf{Rare or subtle visual cues.} Conditions such as snow, temporary detours, and speed cameras were either absent or too small to detect reliably at the given image resolution, resulting in hallucinated or missed detections.
\item \textbf{Fine-grained infrastructure.} Elements like pedestrian refuge islands or painted turning lanes were sometimes misidentified, likely due to resolution constraints in the panoramic image.
\end{itemize}

\paragraph{Key insight: resolution is a limiting factor.}
Many of the observed errors, especially those involving subtle signage, infrastructure details, or rare objects, can be traced directly to insufficient resolution in the source image. While panoramic images provide comprehensive spatial coverage, they often downsample visual detail, making it hard for even a capable MLLM to reliably detect small features such as traffic cones, refuge islands, or painted turn arrows. Switching to higher-resolution imagery, zoomed crops, or targeted visual attention modules could address most of these failures with minimal design change.

\paragraph{Impact on model integrity.}
Importantly, none of the outright failures involved the high-salience geometric or pedestrian-safety hypotheses that dominate model attribution and prediction. This helps explain why \model retains high predictive accuracy despite modest error rates in the long tail of hypotheses. The most informative variables are generally those with clear, resolvable visual signals, precisely the ones the MLLM gets right most often.

\begin{table}[h]
\centering
\small
\caption{Detailed audit of hypotheses flagged as \emph{partially correct} or \emph{incorrect} for the case study in Figure~\ref{fig:case_study_img}.}
\label{tab:vqa_errors}
\begin{tabular}{p{0.33\linewidth}p{0.13\linewidth}p{0.45\linewidth}}
\toprule
\textbf{Hypothesis (paraphrased)} & \textbf{Model Answer} & \textbf{Reason for Deviation} \\
\midrule
\multicolumn{3}{l}{\textit{Partially correct}}\\
\midrule
Are there trees or shrubs that \emph{might} obstruct visibility? & Yes & Small street trees are present, but they do not noticeably block sight lines; the “obstruction’’ qualifier is subjective.\\
Are any advertisements or billboards visible? & Yes & A store fascia (\emph{Hallmark}) is present, yet it is a storefront sign rather than a driver-facing billboard, making the classification debatable.\\
Are reflective road markers visible? & Yes & Lane arrows could contain reflective paint, but this cannot be verified from the daytime image; confidence is therefore partial.\\[4pt]
\midrule
\multicolumn{3}{l}{\textit{Incorrect}}\\
\midrule
Is there visible construction work on the road or sidewalk? & Yes & No construction activity or equipment is present in the scene.\\
Is there a pedestrian refuge island? & Yes & The intersection lacks any raised or painted refuge island.\\
Are pedestrian signals present at the crosswalk? & No & Standard pedestrian countdown signals are visible on the far-left mast arm.\\
Is there a dedicated turning lane? & No & Painted turn arrows are clearly marked in the foreground lane.\\
Are temporary barriers or detours visible? & Yes & No cones, barriers, or road-closure signs can be observed.\\
\bottomrule
\end{tabular}
\end{table}

\paragraph{Additional validation across SVIs.}
We also performed manual checks on a further random sample of 50 SVIs using the same audit procedure. These secondary inspections showed that the high-salience hypotheses, such as lane markings, crosswalks, and medians, were correctly identified in over 90 percent of cases, while most errors continued to involve rare or fine-grained features. This consistency across multiple street views demonstrates that the key variables driving model performance are broadly reliable and not artifacts of a single image.

\paragraph{Practical refinement strategies.}
To further improve reliability without sacrificing transparency, several targeted interventions can be introduced:
\begin{itemize}[leftmargin=*]
\item \textbf{Prompt rephrasing.} Adding clarifying definitions or thresholds (e.g., “construction work must include cones or equipment”) could help disambiguate borderline cases.
\item \textbf{View augmentation.} Supplementing wide-angle views with higher-resolution zoom-ins or directional crops would boost recognition of small but critical features.
\item \textbf{Response calibration.} Incorporating uncertainty scores or allowing abstention on low-confidence answers could help filter out hallucinated positives.
\end{itemize}

\paragraph{Conclusion.}
This case study confirms that MLLMs can reliably answer structured hypotheses about urban form in most settings, especially when the cues are large, unambiguous, and visually distinct. Remaining errors are interpretable, largely due to either visual resolution or semantic vagueness, and can be systematically mitigated. These findings strengthen confidence in \model’s use of MLLM responses as interpretable, robust inputs for scientific analysis.

\subsection{Discussion}


Our proposed framework \model demonstrates the potential of applying MLLMs for interpretable, automated hypothesis discovery in the context of urban safety analysis. The results show that the generated hypotheses not only match or surpass the predictive power of traditional computer vision models but also provide clear semantic insights aligned with established urban design principles. Moreover, the framework enables the discovery of novel, human-interpretable variables that may have been overlooked in previous literature, offering a scalable approach to data-driven scientific discovery.

This work implicitly builds upon a set of assumptions that reflect a shift in how machine learning can be used for knowledge generation. First, we assume that MLLMs, when queried appropriately, can reliably interpret and respond to natural-language hypotheses about complex visual scenes. This assumption effectively treats the MLLM as a proxy for human visual judgment, capable of semantically parsing urban environments in a consistent and informative way. 
Second, we assume that LLMs possess a rational understanding of societal constructs such as safety, risk, and infrastructure, and can leverage this latent knowledge to propose plausible hypotheses. These assumptions align with a broader philosophical view of foundation models not merely as function approximators, but as \emph{cognitive engines}, tools that can externalize latent human reasoning in scalable and programmable ways.

From this perspective, our framework is more than a predictive pipeline, it is a machine-in-the-loop system for structured discovery. It operationalizes a new epistemic loop: language models propose interpretable, theory-aligned variables; MLLMs extract structured representations from raw perceptual data; and statistical models evaluate and refine the space of explanatory factors. While not infallible, this loop offers a novel approach to bridging data, semantics, and scientific reasoning.
Another key implication of this work is that hypothesis generation and refinement, traditionally limited by expert intuition and manual feature engineering, can be guided by LLMs in a statistically grounded loop. This offers a path toward semi-automated scientific workflows where human and machine jointly explore high-dimensional, unstructured data spaces.

Notably, our reliance on foundation models introduces limitations. The correctness of our results depends on the alignment and reliability of the underlying MLLMs and LLMs. Errors in visual understanding or gaps in commonsense reasoning may lead to spurious or irrelevant hypotheses. Moreover, the iterative nature of our approach, while principled, incurs significant computational overhead due to repeated prompting and inference. However, we believe these constraints are temporary. As foundation models continue to improve in efficiency, alignment, and accessibility, the feasibility of such machine-guided discovery frameworks will continue to grow.

\section{Conclusion}

In this paper, we presented \model, a framework that combines MLLMs with interpretable statistical modeling to automate scientific discovery from urban data. Taking road safety in the Manhattan area as a case study, \model formulates natural-language hypotheses, extracts semantically meaningful embeddings through visual question answering, and evaluates their significance using transparent regression models. Our experiments show that \model outperforms conventional deep learning approaches while uncovering novel, interpretable variables aligned with domain knowledge.

This work demonstrates a new paradigm for scientific discovery in urban research, one that integrates perception, language, and statistical reasoning in a unified pipeline. The generality of \model enables broad applicability to other domains such as walkability, equity, and environmental quality, where unstructured data possesses rich information and model interpretability are central. Future work may extend this approach to dynamic data, integrate causal inference, and benefit from ongoing advances in the alignment and efficiency of foundation models. By rethinking machine learning as a tool for interpretable, data-driven reasoning, \model offers a scalable foundation for MLLM hypothesis-driven urban science and beyond.

\clearpage

\bibliographystyle{apalike}  
\bibliography{arxiv}

\begin{thebibliography}{}

\bibitem[Achiam et~al., 2023]{achiam2023gpt}
Achiam, J., Adler, S., Agarwal, S., Ahmad, L., Akkaya, I., Aleman, F.~L., Almeida, D., Altenschmidt, J., Altman, S., Anadkat, S., et~al. (2023).
\newblock Gpt-4 technical report.
\newblock {\em arXiv preprint arXiv:2303.08774}.

\bibitem[Acuto et~al., 2018]{acuto2018building}
Acuto, M., Parnell, S., and Seto, K.~C. (2018).
\newblock Building a global urban science.
\newblock {\em Nature Sustainability}, 1(1):2--4.

\bibitem[Antol et~al., 2015]{antol2015vqa}
Antol, S., Agrawal, A., Lu, J., Mitchell, M., Batra, D., Zitnick, C.~L., and Parikh, D. (2015).
\newblock Vqa: Visual question answering.
\newblock In {\em Proceedings of the IEEE international conference on computer vision}, pages 2425--2433.

\bibitem[Bang et~al., 2025]{bang2025mobility}
Bang, H., Dave, A., Tzortzoglou, F.~N., Wang, S., and Malikopoulos, A.~A. (2025).
\newblock On mobility equity and the promise of emerging transportation systems.
\newblock {\em IEEE Transactions on Intelligent Transportation Systems}.

\bibitem[Batty, 2024]{batty2024computable}
Batty, M. (2024).
\newblock {\em The computable city: histories, technologies, stories, predictions}.
\newblock MIT Press.

\bibitem[Benara et~al., 2024]{benara2024crafting}
Benara, V., Singh, C., Morris, J.~X., Antonello, R.~J., Stoica, I., Huth, A.~G., and Gao, J. (2024).
\newblock Crafting interpretable embeddings for language neuroscience by asking llms questions.
\newblock {\em Advances in neural information processing systems}, 37:124137.

\bibitem[Biljecki and Ito, 2021]{biljecki2021street}
Biljecki, F. and Ito, K. (2021).
\newblock Street view imagery in urban analytics and gis: A review.
\newblock {\em Landscape and Urban Planning}, 215:104217.

\bibitem[Chen et~al., 2024]{chen2024expanding}
Chen, Z., Wang, W., Cao, Y., Liu, Y., Gao, Z., Cui, E., Zhu, J., Ye, S., Tian, H., Liu, Z., et~al. (2024).
\newblock Expanding performance boundaries of open-source multimodal models with model, data, and test-time scaling.
\newblock {\em arXiv preprint arXiv:2412.05271}.

\bibitem[Contributors, 2023]{2023lmdeploy}
Contributors, L. (2023).
\newblock Lmdeploy: A toolkit for compressing, deploying, and serving llm.
\newblock \url{https://github.com/InternLM/lmdeploy}.

\bibitem[Csill{\'e}ry et~al., 2010]{csillery2010approximate}
Csill{\'e}ry, K., Blum, M.~G., Gaggiotti, O.~E., and Fran{\c{c}}ois, O. (2010).
\newblock Approximate bayesian computation (abc) in practice.
\newblock {\em Trends in ecology \& evolution}, 25(7):410--418.

\bibitem[Dosovitskiy et~al., 2020]{dosovitskiy2020image}
Dosovitskiy, A., Beyer, L., Kolesnikov, A., Weissenborn, D., Zhai, X., Unterthiner, T., Dehghani, M., Minderer, M., Heigold, G., Gelly, S., et~al. (2020).
\newblock An image is worth 16x16 words: Transformers for image recognition at scale.
\newblock {\em arXiv preprint arXiv:2010.11929}.

\bibitem[Ewing and Dumbaugh, 2009]{ewing2009built}
Ewing, R. and Dumbaugh, E. (2009).
\newblock The built environment and traffic safety: a review of empirical evidence.
\newblock {\em Journal of Planning Literature}, 23(4):347--367.

\bibitem[Ewing and Handy, 2009]{ewing2009measuring}
Ewing, R. and Handy, S. (2009).
\newblock Measuring the unmeasurable: Urban design qualities related to walkability.
\newblock {\em Journal of Urban design}, 14(1):65--84.

\bibitem[Gettys and Fisher, 1979]{gettys1979hypothesis}
Gettys, C.~F. and Fisher, S.~D. (1979).
\newblock Hypothesis plausibility and hypothesis generation.
\newblock {\em Organizational behavior and human performance}, 24(1):93--110.

\bibitem[Goodfellow et~al., 2016]{goodfellow2016deep}
Goodfellow, I., Bengio, Y., Courville, A., and Bengio, Y. (2016).
\newblock {\em Deep learning}, volume~1.
\newblock MIT press Cambridge.

\bibitem[Gottweis et~al., 2025]{gottweis2025towards}
Gottweis, J., Weng, W.-H., Daryin, A., Tu, T., Palepu, A., Sirkovic, P., Myaskovsky, A., Weissenberger, F., Rong, K., Tanno, R., et~al. (2025).
\newblock Towards an ai co-scientist.
\newblock {\em arXiv preprint arXiv:2502.18864}.

\bibitem[Guzman and Bocarejo, 2017]{guzman2017urban}
Guzman, L.~A. and Bocarejo, J.~P. (2017).
\newblock Urban form and spatial urban equity in bogota, colombia.
\newblock {\em Transportation research procedia}, 25:4491--4506.

\bibitem[Hall, 2012]{hall2012handbook}
Hall, R. (2012).
\newblock {\em Handbook of transportation science}, volume~23.
\newblock Springer Science \& Business Media.

\bibitem[He et~al., 2016]{he2016deep}
He, K., Zhang, X., Ren, S., and Sun, J. (2016).
\newblock Deep residual learning for image recognition.
\newblock In {\em Proceedings of the IEEE conference on computer vision and pattern recognition}, pages 770--778.

\bibitem[Hou et~al., 2020]{hou2020correlated}
Hou, Q., Huo, X., and Leng, J. (2020).
\newblock A correlated random parameters tobit model to analyze the safety effects and temporal instability of factors affecting crash rates.
\newblock {\em Accident Analysis \& Prevention}, 134:105326.

\bibitem[Hu et~al., 2024]{hu2024does}
Hu, Y., Chen, L., and Zhao, Z. (2024).
\newblock How does street environment affect pedestrian crash risks? a link-level analysis using street view image-based pedestrian exposure measurement.
\newblock {\em Accident Analysis \& Prevention}, 205:107682.

\bibitem[Huang et~al., 2024]{huang2024zero}
Huang, W., Wang, J., and Cong, G. (2024).
\newblock Zero-shot urban function inference with street view images through prompting a pretrained vision-language model.
\newblock {\em International Journal of Geographical Information Science}, 38(7):1414--1442.

\bibitem[Huh et~al., 2024]{huh2024platonic}
Huh, M., Cheung, B., Wang, T., and Isola, P. (2024).
\newblock The platonic representation hypothesis.
\newblock {\em arXiv preprint arXiv:2405.07987}.

\bibitem[Hurst et~al., 2024]{hurst2024gpt}
Hurst, A., Lerer, A., Goucher, A.~P., Perelman, A., Ramesh, A., Clark, A., Ostrow, A., Welihinda, A., Hayes, A., Radford, A., et~al. (2024).
\newblock Gpt-4o system card.
\newblock {\em arXiv preprint arXiv:2410.21276}.

\bibitem[Ignatius et~al., 2024]{ignatius2024digital}
Ignatius, M., Lim, J., Gottkehaskamp, B., Fujiwara, K., Miller, C., and Biljecki, F. (2024).
\newblock Digital twin and wearables unveiling pedestrian comfort dynamics and walkability in cities.
\newblock {\em ISPRS Annals of Photogrammetry, Remote Sensing \& Spatial Information Sciences}, 10.

\bibitem[Ke et~al., 2017]{ke2017lightgbm}
Ke, G., Meng, Q., Finley, T., Wang, T., Chen, W., Ma, W., Ye, Q., and Liu, T.-Y. (2017).
\newblock Lightgbm: A highly efficient gradient boosting decision tree.
\newblock {\em Advances in neural information processing systems}, 30.

\bibitem[Kuang et~al., 2025]{kuang2025natural}
Kuang, J., Shen, Y., Xie, J., Luo, H., Xu, Z., Li, R., Li, Y., Cheng, X., Lin, X., and Han, Y. (2025).
\newblock Natural language understanding and inference with mllm in visual question answering: A survey.
\newblock {\em ACM Computing Surveys}, 57(8):1--36.

\bibitem[Lipton, 2018]{lipton2018mythos}
Lipton, Z.~C. (2018).
\newblock The mythos of model interpretability: In machine learning, the concept of interpretability is both important and slippery.
\newblock {\em Queue}, 16(3):31--57.

\bibitem[Liu et~al., 2023]{liu2023visual}
Liu, H., Li, C., Wu, Q., and Lee, Y.~J. (2023).
\newblock Visual instruction tuning.
\newblock {\em Advances in neural information processing systems}, 36:34892--34916.

\bibitem[Lopez et~al., 2025]{lopez2025enhancing}
Lopez, V., Hoang, L., Martinez-Galindo, M., Fern{\'a}ndez-D{\'\i}az, R., Sbodio, M.~L., Ordonez-Hurtado, R., Zayats, M., Mulligan, N., and Bettencourt-Silva, J. (2025).
\newblock Enhancing foundation models for scientific discovery via multimodal knowledge graph representations.
\newblock {\em Journal of Web Semantics}, 84:100845.

\bibitem[Lundberg and Lee, 2017]{lundberg2017unified}
Lundberg, S.~M. and Lee, S.-I. (2017).
\newblock A unified approach to interpreting model predictions.
\newblock {\em Advances in neural information processing systems}, 30.

\bibitem[Ma and Qian, 2018a]{ma2018estimating}
Ma, W. and Qian, Z.~S. (2018a).
\newblock Estimating multi-year 24/7 origin-destination demand using high-granular multi-source traffic data.
\newblock {\em Transportation Research Part C: Emerging Technologies}, 96:96--121.

\bibitem[Ma and Qian, 2018b]{ma2018statistical}
Ma, W. and Qian, Z.~S. (2018b).
\newblock Statistical inference of probabilistic origin-destination demand using day-to-day traffic data.
\newblock {\em Transportation Research Part C: Emerging Technologies}, 88:227--256.

\bibitem[Majchrowska et~al., 2022]{majchrowska2022deep}
Majchrowska, S., Miko{\l}ajczyk, A., Ferlin, M., Klawikowska, Z., Plantykow, M.~A., Kwasigroch, A., and Majek, K. (2022).
\newblock Deep learning-based waste detection in natural and urban environments.
\newblock {\em Waste Management}, 138:274--284.

\bibitem[Montgomery et~al., 2021]{montgomery2021introduction}
Montgomery, D.~C., Peck, E.~A., and Vining, G.~G. (2021).
\newblock {\em Introduction to linear regression analysis}.
\newblock John Wiley \& Sons.

\bibitem[Moon, 1996]{moon1996expectation}
Moon, T.~K. (1996).
\newblock The expectation-maximization algorithm.
\newblock {\em IEEE Signal processing magazine}, 13(6):47--60.

\bibitem[Naveed et~al., 2023]{naveed2023comprehensive}
Naveed, H., Khan, A.~U., Qiu, S., Saqib, M., Anwar, S., Usman, M., Akhtar, N., Barnes, N., and Mian, A. (2023).
\newblock A comprehensive overview of large language models.
\newblock {\em arXiv preprint arXiv:2307.06435}.

\bibitem[Nie et~al., 2025]{nie2025exploring}
Nie, T., Sun, J., and Ma, W. (2025).
\newblock Exploring the roles of large language models in reshaping transportation systems: A survey, framework, and roadmap.
\newblock {\em arXiv preprint arXiv:2503.21411}.

\bibitem[Radford et~al., 2021]{radford2021learning}
Radford, A., Kim, J.~W., Hallacy, C., Ramesh, A., Goh, G., Agarwal, S., Sastry, G., Askell, A., Mishkin, P., Clark, J., et~al. (2021).
\newblock Learning transferable visual models from natural language supervision.
\newblock In {\em International conference on machine learning}, pages 8748--8763. PmLR.

\bibitem[Ruder, 2016]{ruder2016overview}
Ruder, S. (2016).
\newblock An overview of gradient descent optimization algorithms.
\newblock {\em arXiv preprint arXiv:1609.04747}.

\bibitem[Santamouris, 2013]{santamouris2013energy}
Santamouris, M. (2013).
\newblock {\em Energy and climate in the urban built environment}.
\newblock Routledge.

\bibitem[Sunn{\aa}ker et~al., 2013]{sunnaaker2013approximate}
Sunn{\aa}ker, M., Busetto, A.~G., Numminen, E., Corander, J., Foll, M., and Dessimoz, C. (2013).
\newblock Approximate bayesian computation.
\newblock {\em PLoS computational biology}, 9(1):e1002803.

\bibitem[Tang et~al., 2025]{tang2025large}
Tang, Y., Kong, M., and Sun, L. (2025).
\newblock Large language models for data synthesis.
\newblock {\em arXiv preprint arXiv:2505.14752}.

\bibitem[Tang et~al., 2024]{tang2024itinera}
Tang, Y., Wang, Z., Qu, A., Yan, Y., Wu, Z., Zhuang, D., Kai, J., Hou, K., Guo, X., Zhao, J., et~al. (2024).
\newblock Itinera: Integrating spatial optimization with large language models for open-domain urban itinerary planning.
\newblock In {\em Proceedings of the 2024 Conference on Empirical Methods in Natural Language Processing: Industry Track}, pages 1413--1432.

\bibitem[Thawakar et~al., 2025]{thawakar2025llamav}
Thawakar, O., Dissanayake, D., More, K., Thawkar, R., Heakl, A., Ahsan, N., Li, Y., Zumri, M., Lahoud, J., Anwer, R.~M., et~al. (2025).
\newblock Llamav-o1: Rethinking step-by-step visual reasoning in llms.
\newblock {\em arXiv preprint arXiv:2501.06186}.

\bibitem[Wang et~al., 2025]{wang2025parameter}
Wang, Z., Zhu, X., Yang, X., Luo, G., Li, H., Tian, C., Dou, W., Ge, J., Lu, L., Qiao, Y., et~al. (2025).
\newblock Parameter-inverted image pyramid networks for visual perception and multimodal understanding.
\newblock {\em arXiv preprint arXiv:2501.07783}.

\bibitem[Wei et~al., 2022a]{wei2022emergent}
Wei, J., Tay, Y., Bommasani, R., Raffel, C., Zoph, B., Borgeaud, S., Yogatama, D., Bosma, M., Zhou, D., Metzler, D., et~al. (2022a).
\newblock Emergent abilities of large language models.
\newblock {\em arXiv preprint arXiv:2206.07682}.

\bibitem[Wei et~al., 2022b]{wei2022chain}
Wei, J., Wang, X., Schuurmans, D., Bosma, M., Xia, F., Chi, E., Le, Q.~V., Zhou, D., et~al. (2022b).
\newblock Chain-of-thought prompting elicits reasoning in large language models.
\newblock {\em Advances in neural information processing systems}, 35:24824--24837.

\bibitem[Wong and Lee, 2005]{wong2005statistical}
Wong, W. and Lee, J. (2005).
\newblock {\em Statistical analysis of geographic information with ArcView GIS and ArcGIS}.
\newblock Wiley.

\bibitem[Wu et~al., 2023]{wu2023multimodal}
Wu, J., Gan, W., Chen, Z., Wan, S., and Yu, P.~S. (2023).
\newblock Multimodal large language models: A survey.
\newblock In {\em 2023 IEEE International Conference on Big Data (BigData)}, pages 2247--2256. IEEE.

\bibitem[Xia et~al., 2025]{xia2025reimagining}
Xia, Y., Qu, A., Zheng, Y., Tang, Y., Zhuang, D., Liang, Y., Wang, S., Wu, C., Sun, L., Zimmermann, R., and Zhao, J. (2025).
\newblock Reimagining urban science: Scaling causal inference with large language models.
\newblock {\em arXiv preprint arXiv:2504.12345}.

\bibitem[Xue et~al., 2024]{xue2024integrating}
Xue, H., Guo, P., Li, Y., and Ma, J. (2024).
\newblock Integrating visual factors in crash rate analysis at intersections: An automl and shap approach towards cycling safety.
\newblock {\em Accident Analysis \& Prevention}, 200:107544.

\bibitem[Yang et~al., 2023]{yang2023dawn}
Yang, Z., Li, L., Lin, K., Wang, J., Lin, C.-C., Liu, Z., and Wang, L. (2023).
\newblock The dawn of lmms: Preliminary explorations with gpt-4v (ision).
\newblock {\em arXiv preprint arXiv:2309.17421}, 9(1):1.

\bibitem[Yu et~al., 2024]{yu2024can}
Yu, X., Ma, J., Tang, Y., Yang, T., and Jiang, F. (2024).
\newblock Can we trust our eyes? interpreting the misperception of road safety from street view images and deep learning.
\newblock {\em Accident Analysis \& Prevention}, 197:107455.

\bibitem[Yue et~al., 2025]{yue2025does}
Yue, Y., Chen, Z., Lu, R., Zhao, A., Wang, Z., Song, S., and Huang, G. (2025).
\newblock Does reinforcement learning really incentivize reasoning capacity in llms beyond the base model?
\newblock {\em arXiv preprint arXiv:2504.13837}.

\bibitem[Zeng et~al., 2017]{zeng2017multivariate}
Zeng, Q., Wen, H., Huang, H., Pei, X., and Wong, S. (2017).
\newblock A multivariate random-parameters tobit model for analyzing highway crash rates by injury severity.
\newblock {\em Accident Analysis \& Prevention}, 99:184--191.

\bibitem[Zhang et~al., 2024]{zhang2024mm}
Zhang, D., Yu, Y., Dong, J., Li, C., Su, D., Chu, C., and Yu, D. (2024).
\newblock Mm-llms: Recent advances in multimodal large language models.
\newblock {\em arXiv preprint arXiv:2401.13601}.

\end{thebibliography}

\clearpage
\appendix
\section*{Appendix}

\section{Prompt Design and Sampling Strategy} 
\label{appendix:prompt}

\subsection{Hypothesis Prompting Strategy.}

In our implementation, we adopt an exploration-exploitation strategy for generating new hypotheses using Large Language Models. At each iteration $t$, a new set of candidate hypotheses $\mathcal{H}^t$ is sampled by prompting the LLM with one of two designed templates:

\vspace{5pt}
\begin{tcolorbox}[title=HYPO\_EXPLOIT\_PROMPT, breakable]
\begin{minted}{markdown}
Generate a new set of {n} diverse and non-overlapping binary classification questions that capture interpretable features from street view images relevant to predicting crash rates.

Existing questions: ```{hypo t-1}```

Design requirements:
- Avoid redundancy and ensure each question is unique.
- Prioritize features with clear relevance to road safety.
- Questions must be answerable based solely on the visual content of the street view image.
- Each question must have exactly two answer choices.
- The answer options must be mutually exclusive and collectively exhaustive.
- Avoid vague, ambiguous, or overly subjective formulations.
- Focus on concrete visual cues (e.g., road structure, visibility, signage, obstructions) that could improve crash prediction.

First, provide a short reasoning paragraph analyzing the strengths and weaknesses of the existing questions with respect to safety prediction. Suggest new aspects that could be captured to improve performance.

Then return a valid Python dictionary formatted as JSON:
{{
    "reasoning": "short reasoning text.",
    "Question 1": {{"0": "Answer option A", "1": "Answer option B"}},
    ...
}}

The response must be valid JSON (parseable by `json.loads`) and should include only the JSON dictionary, without additional commentary.
\end{minted}
\end{tcolorbox}
\vspace{5pt}

An exploitation prompt (see HYPO\_EXPLOIT\_PROMPT) that conditions on the currently retained hypothesis set $\mathcal{H}^{t-1}$ and their statistical significance $\mathcal{P}^{t-1}$. This prompt encourages refinement and expansion of hypotheses with known predictive value, anchoring the generation process to empirically validated ideas.

\vspace{5pt}
\begin{tcolorbox}[title=HYPO\_EXPLORE\_PROMPT, breakable]
\begin{minted}{markdown}
Generate {n} unique binary classification questions that explore the broad visual, social, and contextual dimensions of street view imagery.

Design requirements:
- Avoid repetition and ensure semantic diversity.
- Questions must be clearly answerable from a single street view image.
- Each question must have exactly two answer choices.
- The answer options must be mutually exclusive and collectively exhaustive.
- Avoid overly ambiguous, speculative, or subjective content.

Consider features beyond the obvious, such as socioeconomic signals, environmental quality, cultural context, artistic elements, or neighborhood character. Let curiosity guide you to ask unconventional, thought-provoking questions that reveal hidden or surprising correlations.

Return a valid Python dictionary formatted as JSON:
{{
    "reasoning": "brief explanation of how the questions were conceived.",
    "Question 1": {{"0": "Answer option A", "1": "Answer option B"}},
    ...
}}

The response must be valid JSON (parseable by `json.loads`) and should include only the JSON dictionary, without additional commentary.
\end{minted}
\end{tcolorbox}
\vspace{5pt}

An exploration prompt (see HYPO\_EXPLORE\_PROMPT) that deliberately encourages broader semantic coverage, open-ended question generation, and inclusion of unconventional or underexplored visual features. This promotes diversity and mitigates local optima.

To balance these goals, we apply a stochastic control mechanism: at each iteration, with a fixed probability $p_{\text{explore}}$ (default 0.1), the exploration prompt is selected; otherwise, the exploitation prompt is used. This simple sampling scheme mirrors common strategies in reinforcement learning and approximate inference, ensuring both local exploitation and global search over the hypothesis space.

The final prompt used in each iteration is automatically constructed based on the current retained hypotheses and plugged into the appropriate template. Each LLM response is parsed as structured JSON and incorporated into the next round of variable construction and statistical evaluation.

This approach enables interpretable and data-driven exploration of the hypothesis space while maintaining relevance and control through prior feedback and statistical validation.

\subsection{Embedding Prompting Strategy}

For embedding construction, we use Multimodal Large Language Models (MLLMs) to answer each generated hypothesis based on the visual content of a street view image. 
We define two templates for prompting the MLLM: a single-question version for sequential evaluation, and a batched version for more efficient parallel processing.

\vspace{5pt}
\begin{tcolorbox}[title=EMB\_PROMPT, breakable]
\begin{minted}{markdown}
Based on the street view image of a road, answer the following question: 
```{hypo t}```

Return only an integer indicating the option number and nothing else
\end{minted}
\end{tcolorbox}
\vspace{5pt}

The single-question prompt (EMB\_PROMPT) is used to infer the answer to one hypothesis at a time. It specifies the question and options, instructing the MLLM to return only the index of the chosen option as an integer.

\vspace{5pt}
\begin{tcolorbox}[title=EMB\_BATCH\_PROMPT, breakable]
\begin{minted}{markdown}
Based on the street view image of a road, answer the following questions: 
```{hypo t}```

example response format (should be replaced by your answer to the questions):
["0", "1", ...]

Return only a list of {n} integers indicating the option numbers, the response should be parsable with `json.loads` and nothing else
\end{minted}
\end{tcolorbox}
\vspace{5pt}

The batch prompt (EMB\_BATCH\_PROMPT) enables simultaneous evaluation of multiple hypotheses. It presents all questions at once and expects a list of integers corresponding to the chosen answer for each.
Unless otherwise specified, we use the batch prompt for all experiments, as we find that it does not compromise VQA quality.
This structured prompting strategy ensures semantic traceability, data efficiency, and ease of downstream statistical modeling.

\clearpage

\section{Summary of Built Environment Variables}
\label{appx:bvar}

This section summarizes the 58 built environment variables used in the study. These variables are derived from multiple data sources, including Google Street View images, New York City open data, land use polygons, OpenStreetMap, and POI (Point of Interest) datasets. The variables are categorized into five main classes: View Indices, Road Attributes, Land Use, Points of Interest (POI), and Traffic-related Facilities.

\renewcommand{\arraystretch}{.9}
\begin{table}[ht]
\centering
\caption{Summary of Built Environment Variables}
\label{tab:built_env_vars}
\resizebox{.96\linewidth}{!}{
\begin{tabular}{@{}lp{8cm}p{4.5cm}@{}}
\toprule
\textbf{Category} & \textbf{Variables} & \textbf{Description} \\
\midrule

\textbf{View Indices (12)} & 
\texttt{road\_view\_index, pavement\_view\_index,} \newline
\texttt{sky\_view\_index, building\_view\_index,} \newline
\texttt{tree\_view\_index, grass\_view\_index,} \newline
\texttt{fence\_view\_index, wall\_view\_index,} \newline
\texttt{traffic\_lights\_area\_view\_index, stop\_signs\_area\_view\_index,} \newline
\texttt{traffic\_lights\_number\_view\_index, stop\_signs\_number\_view\_index} & 
Proportion of image pixels representing each element (e.g., sky, pavement, building, vegetation, traffic signs), extracted from Google Street View images. \\

\addlinespace[0.5em]
\textbf{Road Attributes (2)} & 
\texttt{Highway\_or\_not, st\_width} & 
Binary indicator of whether the road is a highway, and the width of the road in meters. \\

\addlinespace[0.5em]
\textbf{Land Use (7)} & 
\texttt{Residential\_land, Commercial\_land,} \newline
\texttt{Industrial\_land, Transportation\_land,} \newline
\texttt{Public\_land, Open\_space, Land\_use\_mixture} & 
Area of six land use types within a buffer around the road segment, and a land use mixture index capturing the diversity of land use types. \\

\addlinespace[0.5em]
\textbf{POI (27)} & 
\texttt{number\_500\_Residential,} \newline
\texttt{number\_500\_Education\_Facility, number\_500\_Cultural\_Facility,} \newline
\texttt{number\_500\_Recreational\_Facility, number\_500\_Social\_Services, number\_500\_Transportation\_Facility,} \newline
\texttt{number\_500\_Commercial,} \newline
\texttt{number\_500\_Government\_Facility,} \newline
\texttt{number\_500\_Religious\_Institution,} \newline
\texttt{number\_500\_Health\_Services,} \newline
\texttt{number\_500\_Public\_Safety,} \newline
\texttt{number\_500\_Water, number\_500\_Miscellaneous,} \newline
\texttt{distance\_Residential, distance\_Education\_Facility, distance\_Cultural\_Facility,} \newline
\texttt{distance\_Recreational\_Facility, distance\_Social\_Services,} \newline
\texttt{distance\_Transportation\_Facility,} \newline
\texttt{distance\_Commercial, distance\_Government\_Facility, distance\_Religious\_Institution,} \newline
\texttt{distance\_Health\_Services, distance\_Public\_Safety, distance\_Water, distance\_Miscellaneous, POI\_type} & 
Number of POIs of 13 types within a 500-meter buffer, distances to the nearest POI of each type, and the primary POI type. \\

\addlinespace[0.5em]
\textbf{Traffic-related Facilities (10)} & 
\texttt{number\_500\_transport\_bus\_stop, number\_500\_traffic\_stop, number\_500\_traffic\_crossing,} \newline
\texttt{number\_500\_traffic\_motorway\_junction, number\_500\_traffic\_traffic\_signals,} \newline
\texttt{distance\_transport\_bus\_stop, distance\_traffic\_stop, distance\_traffic\_crossing,} \newline
\texttt{distance\_traffic\_motorway\_junction, distance\_traffic\_traffic\_signals} & 
Number and distance of various traffic-related facilities (e.g., bus stops, crossings, traffic lights, junctions) within 500 meters of the road segment. \\

\bottomrule
\end{tabular}
}
\end{table}

\clearpage

\begin{figure}[h]
    \centering
    \includegraphics[width=.98\linewidth]{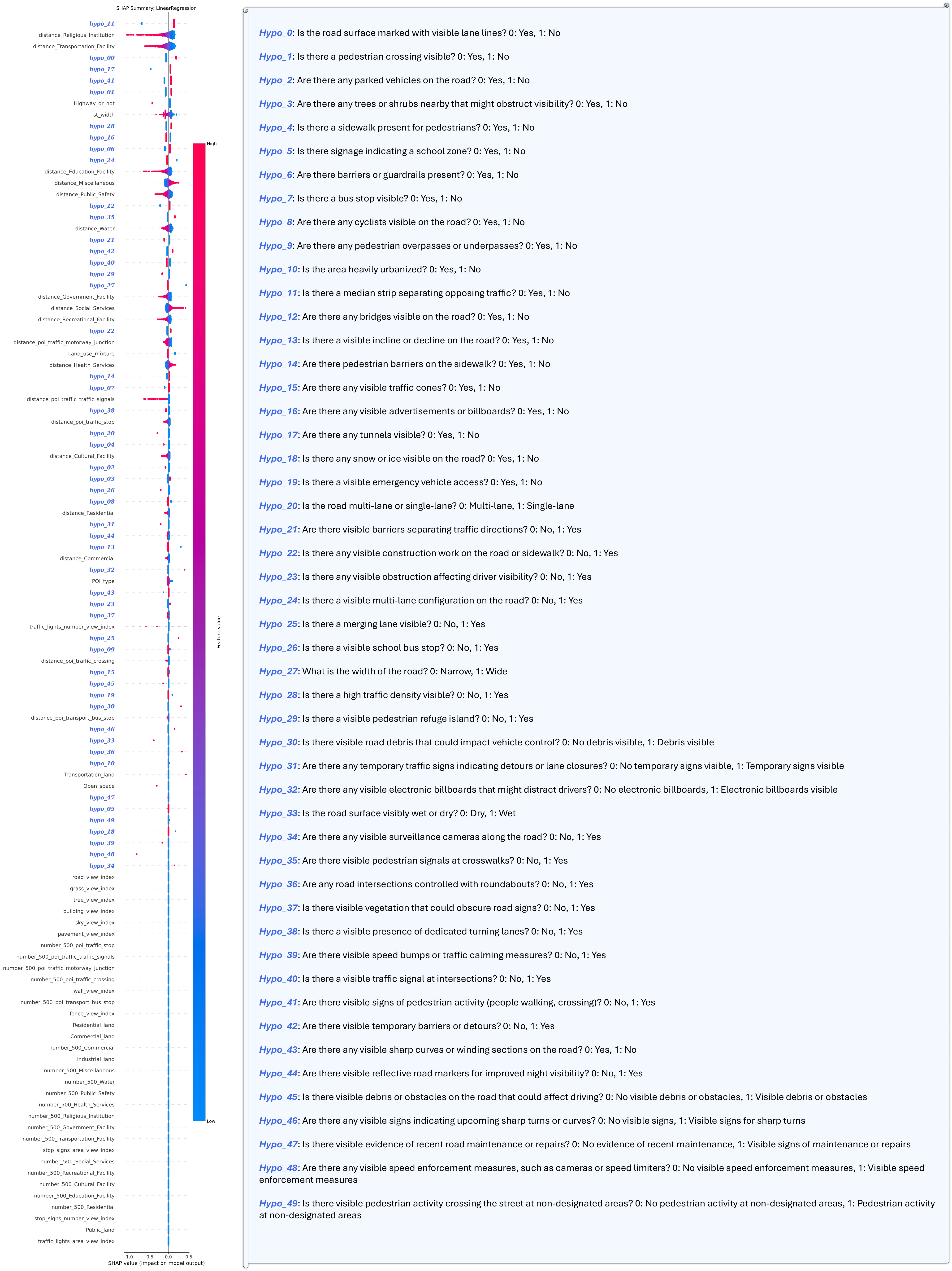}
    \caption{Final set of 50 natural-language hypotheses retained by \model after the iterative process.}
    \label{fig:full_hypo}
\end{figure}

\clearpage

 \section{Qualitative Analysis of the Final Hypothesis Set}
\label{appendix:hypo_analysis}

The final iteration of \model retains 50 hypotheses that together provide a multifaceted description of the street environment, the resulting hypotheses are shown in Figure~\ref{fig:full_hypo}.  A close reading of the list reveals several encouraging patterns as well as a few limitations that suggest directions for future refinement.

\vspace{0.5em}
\noindent\textbf{Breadth of physical design cues.}  
A substantial portion of the questions targets canonical elements of roadway geometry: median strips, lane markings, multi-lane configurations, and road width.  These queries parallel variables that civil engineers traditionally collect through manual audits, yet here they arise automatically from the LLM without prior codification.  Their presence confirms that the system can rediscover core safety factors in a purely data-driven manner.

\vspace{0.5em}
\noindent\textbf{Attention to vulnerable users.}  
Another cluster of hypotheses concentrates on pedestrian infrastructure and activity, including crossings, signals, overpasses, and informal crossing behaviour.  The model further probes cyclist visibility and the existence of school-zone signage.  By elevating these human-centric factors to high SHAP ranks, the framework highlights elements often under-represented in purely geometric crash models, reinforcing its potential for policy-relevant discovery.

\vspace{0.5em}
\noindent\textbf{Contextual and perceptual variables.}  
Several questions extend beyond traditional inventories to capture visual distraction (billboards, electronic signage), driver visibility (obstructions, vegetation), and transient obstacles (construction work, debris, snow or ice).  Such perceptual cues are rarely present in structured GIS layers yet can be critical for real-world safety.  Their emergence illustrates how image-based hypothesis generation can broaden the discourse on urban risk.

\vspace{0.5em}
\noindent\textbf{Redundancy and granularity considerations.}  
A few hypotheses partly overlap in meaning.  For instance, both the multi-lane question and the lane-width question address capacity, and both barrier-related items refer to physical separation.  While some redundancy is expected in an open-ended search, it suggests an opportunity to introduce a post-generation clustering step that merges semantically similar queries and thereby yields a more parsimonious variable set.

\vspace{0.5em}
\noindent\textbf{Ambiguity and data-imbalance.}  
Certain questions may be ambiguous in practice or rarely triggered in the Manhattan data.  Examples include snow or ice on the roadway and visible emergency vehicle access, which occur infrequently in the imagery and hence contribute little statistical signal.  Future iterations might incorporate an adaptive pruning rule that removes hypotheses with low prevalence or high annotation uncertainty.

\vspace{0.5em}
\noindent\textbf{Categorical framing limitations.}  
All queries are currently coded as questions with categorical answers.  While this choice simplifies MLLM inference and statistical testing, it can obscure gradations of exposure.  Road width, traffic density, and billboard prominence are inherently continuous or ordinal.  Extending \model to multi-class or scalar responses would permit richer descriptions while retaining interpretability.

\vspace{0.5em}
\noindent\textbf{Implications for urban research.}  
Taken together, the hypothesis set demonstrates that \model can surface both established and novel factors spanning geometry, infrastructure, human activity, and perceptual context.  Even with the noted redundancies and simplifications, the variables provide a transparent basis for scientific inquiry, enabling planners to trace crash-risk patterns back to concrete visual attributes.  As MLLMs continue to improve, we anticipate the framework will yield even finer-grained hypotheses, opening new avenues for theory building in urban safety and beyond.

\end{document}